\pdfoutput=1

\documentclass[11pt]{article}

\usepackage[preprint]{acl}

\usepackage{times}
\usepackage{latexsym}

\usepackage[T1]{fontenc}

\usepackage[utf8]{inputenc}

\usepackage{microtype}

\usepackage{inconsolata}

\usepackage{hyperref}       
\usepackage{url}            
\usepackage{booktabs}       
\usepackage{amsfonts}       
\usepackage{graphicx}       
\usepackage{nicefrac}       
\usepackage{microtype}      
\usepackage{xcolor}         
\usepackage{longtable}
\usepackage{booktabs}
\usepackage{multirow}
\usepackage{amsmath}
\usepackage{tabularx}
\usepackage{listings}
\usepackage{lscape}
\usepackage{enumitem}
\usepackage{subcaption}

\newcommand\datasetname{\textcolor{black}{\textsc{SEADialogues}}}

\usepackage{float}

\title{$\datasetname$: A Multilingual Culturally Grounded Multi-turn Dialogue Dataset on Southeast Asian Languages}

\author{Muhammad Dehan Al Kautsar$^{1}\thanks{The authors contributed equally.}$, Aswin Candra$^{2*}$, Muhammad Alif Al Hakim$^{3*}$,\\
\textbf{Maxalmina Satria Kahfi$^{4*}$, Fajri Koto$^1$, Alham Fikri Aji$^1$, Peerat Limkonchotiwat$^5$,}\\ 
\textbf{Ekapol Chuangsuwanich$^6$, Genta Indra Winata$^7$}\\
  $^1$MBZUAI$\quad$$^2$Mekari$\quad$$^3$Universitas Indonesia$\quad$$^4$Detik Network$\quad$$^5$AI Singapore\\
  $^6$Chulalongkorn University$\quad$$^7$Capital One\\
  \texttt{muhammad.dehan@mbzuai.ac.ae}}

\begin{document}
\maketitle
\begin{abstract}
Although numerous datasets have been developed to support dialogue systems, most existing chit-chat datasets overlook the cultural nuances inherent in natural human conversations. To address this gap, we introduce $\datasetname$, a culturally grounded dialogue dataset centered on Southeast Asia, a region with over 700 million people and immense cultural diversity. Our dataset features dialogues in eight languages from six Southeast Asian countries, many of which are low-resource despite having sizable speaker populations. To enhance cultural relevance and personalization, each dialogue includes persona attributes and two culturally grounded topics that reflect everyday life in the respective communities. Furthermore, we release a multi-turn dialogue dataset to advance research on culturally aware and human-centric large language models, including conversational dialogue agents.
\end{abstract}

\section{Introduction}

Dialogue systems have made significant strides in enabling real-life interactions, from task-oriented models that assist users with specific goals, such as booking flights or restaurants~\cite{budzianowski2018multiwoz,chakraborty2025t1} or managing schedules~\cite{mo2024hiertod}, to chit-chat systems designed for more casual, extended conversations~\cite{lin2021xpersona,sun2021adding}. Although a wide range of datasets exist, particularly for open-domain dialogue, most were not created with cultural sensitivity in mind and therefore fail to capture the nuanced ways in which culture shapes human communication. While large language models (LLMs) have substantially advanced the development of dialogue systems, their direct use often makes them struggle to accurately reflect cultural values, particularly when generating culture-specific references or contextually grounded entities~\cite{adilazuarda2024towards, chiu2024culturalbench}. This limitation highlights the need for culturally enriched dialogue datasets that have contextual relevance in real-world applications. 

\begin{figure}
    \centering
    \small
    \includegraphics[width=\linewidth]{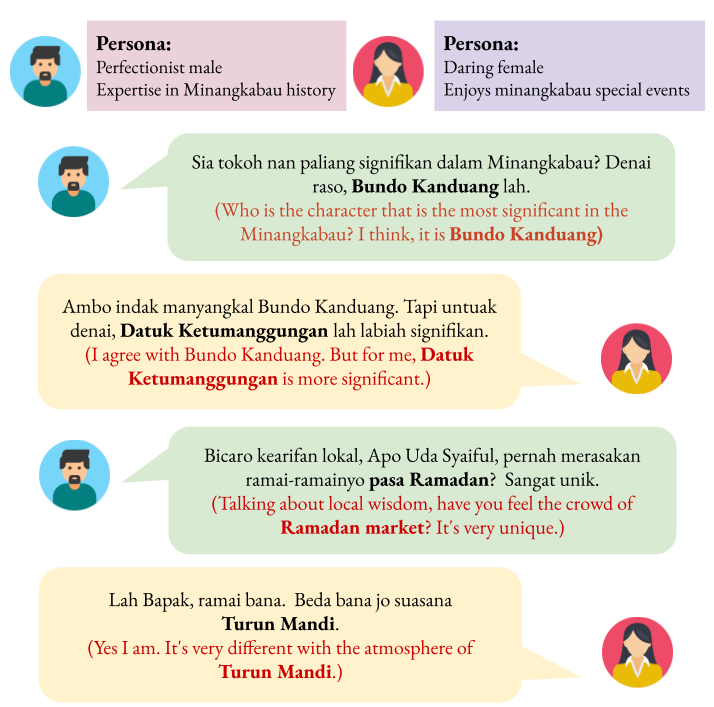}
    \caption{Example dialogue between two individuals, with personas incorporated to ensure the conversation reflects their distinct characteristics.}
    \label{fig:dialogue_example}
    \vspace{-3mm}
\end{figure}

Previous work has demonstrated that incorporating local entities and leveraging multilingual data augmentation can significantly enhance performance and improve the cultural awareness of LLMs~\cite{ding2022globalwoz}. Additionally, modeling user personas has been shown to increase the naturalness and engagement of dialogue systems, enabling more personalized and human-like interactions~\cite{zhang2018personalizing}. In the context of Southeast Asia, home to over 700 million people across 11 countries,\footnote{\url{https://www.worldometers.info/world-population/south-eastern-asia-population/}} only a limited number of languages have been explored in dialogue system research. Most existing work has focused on languages such as Indonesian~\cite{lin2021xpersona,kautsar2023indotod}, Thai~\cite{robloke2019task}, and Vietnamese~\cite{van2022viwoz}, yet key challenges remain due to the region's linguistic diversity and deep cultural influences~\cite{aji2022one}. Foremost among these challenges are the scarcity of large-scale annotated dialogue datasets and an overreliance on translated English corpora. Such translations often produce unnatural conversational flows that fail to capture the cultural and linguistic nuances of the target languages, even when local entities are included. Additionally, research on synthetic versus culturally-grounded multi-turn dialogue generation using LLMs, particularly in the context of the Global South, also remains largely overlooked. Existing datasets~\cite{zhang2018personalizing, ding2022globalwoz, ding2023enhancing} continue to focus predominantly on the Global North and often fail to capture the cultural nuances present in both human-human and human-machine interactions.

To address the aforementioned challenges, we propose $\datasetname$, a benchmark dataset featuring multi-turn, culturally grounded, and persona-rich conversations in 8 languages across 6 Southeast Asian countries: Indonesian, Javanese, Minangkabau, Thai, Malay, Vietnamese, Tamil, and Tagalog. It consists of 32,000 dialogues covering over 100 culturally relevant topics, with each dialogue addressing multiple topics, as shown in Figure~\ref{fig:dialogue_example}. It also includes 210 diverse personas to support personalized and culturally aware dialogue generation. Our detailed dataset statistics are explained in Appendix~\ref{sec:dataset_statistics}.

Our contributions can be summarized in the following aspects:

\begin{itemize}
    \item We introduce $\datasetname$,\footnote{The dataset can be accessed at~\url{https://huggingface.co/datasets/SEACrowd/SEADialogues}, and the corresponding code is publicly available at~\url{https://github.com/SEACrowd/SEADialogues}.} a new open-source, multilingual, multi-turn, and persona-rich synthetic dialogue dataset encompassing eight languages across six Southeast Asian countries. The dialogues are LLM-generated and are carefully tailored to reflect local cultures, values, and region-specific topics. Additionally, we construct culturally grounded personas from each country to ensure realistic and contextually accurate interactions.
    \item We study the effectiveness of generating synthetic multi-turn SEA dialogues with open-weights and proprietary LLMs, assessing their ability to produce culturally appropriate and persona-consistent dialogue.
    \item We study the correlation between human annotations and LLM judges across different aspects and metrics, finding that G-Eval with the GPT-4.1 mini model exhibits a good correlation with human annotations. However, LLM judges still require improvement to match human evaluations for SEA dialogues.
\end{itemize}

\begin{figure*}
    \centering
    \includegraphics[width=\linewidth]{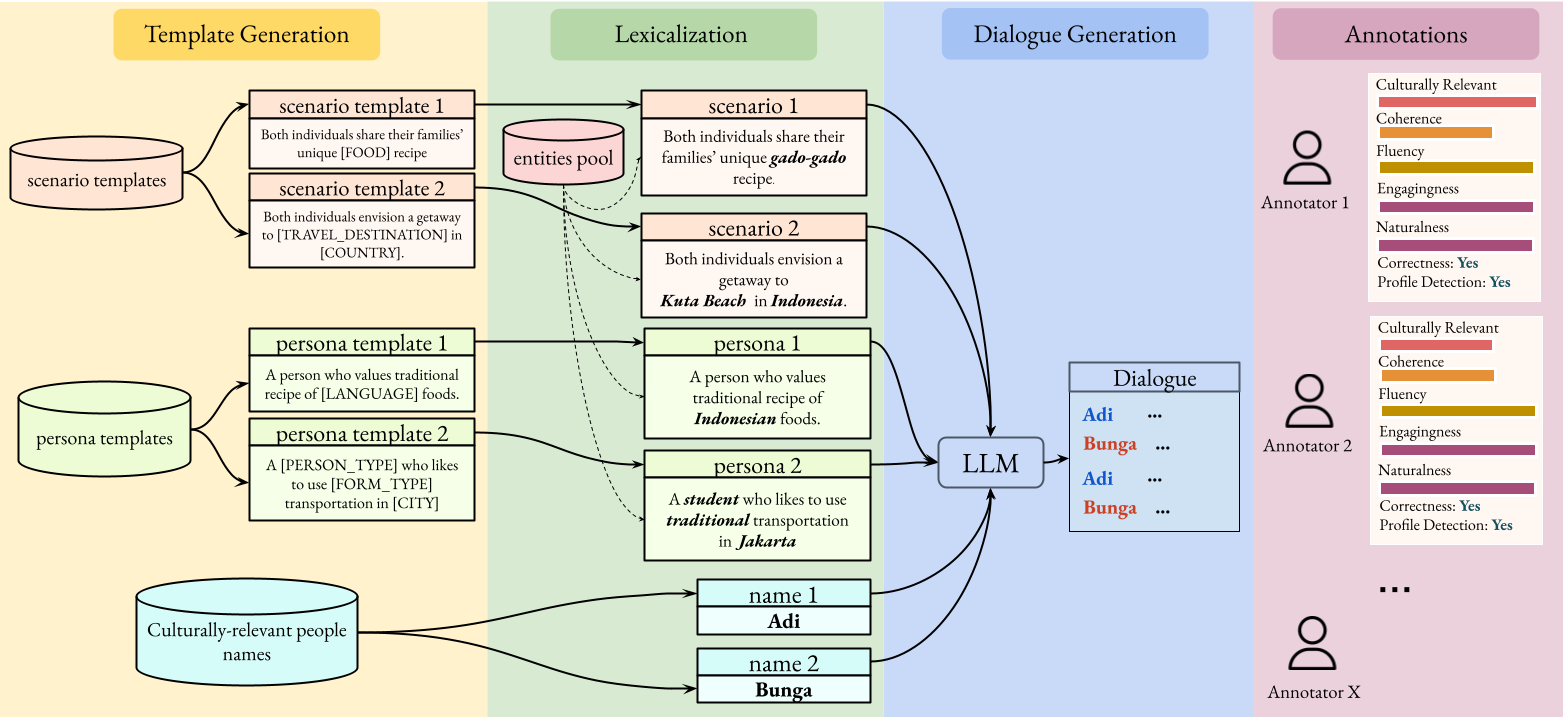}
    \caption{Overview of the $\datasetname$ generation pipeline, which comprises four key stages: (1) template generation, (2) lexicalization of cultural elements, (3) synthetic dialogue generation using LLMs, and (4) final annotation by human annotators. In the first stage, templates with delexicalized entities are created; these placeholders are then populated with culturally relevance entities during lexicalization. Next, multi-turn dialogues are generated synthetically based on speaker personas. Finally, human annotators evaluate the dialogues using quality metrics.}
    \label{fig:pipeline}
\end{figure*}

\section{What Factors Make a Good Dataset?}
\paragraph{Culturally-Relevant Information or Entities.}
Translation-based dialogue datasets often directly translate English named entities (e.g., hotels, locations), even when such entities are culturally irrelevant or nonexistent in the target regions, leading to unnatural and impractical interactions \cite{ding2022globalwoz}. For instance, a system targeting users in Dubai may inappropriately reference Cambridge-specific entities or postcode systems. \citet{hu2023multi} attempt to mitigate this through cultural adaptation strategies such as entity replacement and value redistribution across cities like Dubai, Paris, and Ankara. However, these methods often overlook deeper cultural nuances, regional language variation, and communication styles. As dialogue systems are increasingly deployed in diverse sociocultural settings, grounding them in culturally relevant knowledge is essential for generating coherent, relatable, and effective conversations.

\paragraph{Is LLM enough to generate good data?}
LLMs have advanced the generation of high-quality synthetic data, addressing challenges related to data scarcity and privacy. Their capacity to produce contextually rich, human-like text supports the creation of training datasets across diverse domains, including healthcare \cite{peng2023study} and education \cite{moore2023empowering}. Recent efforts in dialogue data generation leverage different LLMs and few-shot examples from datasets like DialogSum and SAMSum to synthesize dialogue data \cite{suresh2025diasynth}. However, the generation process can suffer from limited domain knowledge, particularly on topics underrepresented in the models’ pretraining data. This issue becomes especially pronounced when generating personas and subtopics using zero-shot prompting.

While LLMs implicitly encode a vast amount of cultural knowledge, prior studies have shown that they often fail to apply this knowledge effectively in context. Explicitly providing relevant cultural information has been shown to significantly enhance the specificity, cultural sensitivity, and overall quality of responses in intercultural dialogue tasks \cite{nguyen2024cultural}. These findings suggest that even advanced models benefit from structured and cultural knowledge during data generation.

\paragraph{How does our dataset differ from existing datasets?}
To address the limitations of prior translation-based and LLM-generated dialogue datasets in representing cultural and local relevance, we propose a multilingual synthetic dialogue dataset covering eight languages across six Southeast Asian countries. 
Unlike existing approaches that rely solely on zero-shot prompting or entity replacement, our dataset integrates manually curated cultural knowledge, such as local entities, food, and communication norms, directly into the prompt design. Using LLMs, we generate dialogues conditioned on user personas, dialogue topics, and region-specific cultural contexts, aiming to produce coherent, culturally grounded conversations that reflect real-world user behavior. This approach bridges the gap between linguistic diversity and cultural representation in dialogue systems, supporting more inclusive and contextually aware conversational agents for underrepresented regions.

In addition to human evaluation, we propose the use of LLMs as automated judges to assess multi-turn dialogues along dimensions such as coherence, fluency, and cultural relevance. To our knowledge, this is the first dialogue dataset evaluated by LLMs for multi-turn conversations in a multilingual, culturally grounded setting. This dual evaluation framework enhances scalability while maintaining rigor, supporting the development of more inclusive and context-aware dialogue systems for underrepresented regions.

\begin{table*}[!ht]
\centering
\resizebox{\textwidth}{!}{
\begin{tabular}{l|cccccccc}
\toprule
\textbf{Dataset} & \textbf{\#Lang.} & \textbf{\#Dial.} & \textbf{Topic Type} & \textbf{\#Scenario} & \textbf{\#Persona} & \textbf{Human Annotations} & \textbf{$\neg$Translation} & \textbf{Cultural Rel.} \\
\midrule
DailyDialog~\cite{li2017dailydialog} & 1 & 13.1K & Single & 10 & 0 & $\checkmark$ & $\checkmark$ & $\times$ \\
MultiWOZ \cite{budzianowski2018multiwoz} & 1 & 8.4K  & Single & - & - & $\checkmark$ & $\times$ & $\times$ \\
PERSONA-CHAT \cite{zhang2018personalizing} & 1 & 10.9K & Multiple & - & 1,155 & $\checkmark$ & $\checkmark$ & $\times$ \\ 
PersonalDialog \cite{zheng2019personalized} & 1 & 20.83M  & Single & - & - & $\checkmark$ & $\checkmark$ & $\times$ \\ 
CrossWOZ \cite{zhu2020crosswoz} & 1 & 5K  & Single & - & - & $\checkmark$ & $\checkmark$ & $\times$ \\
MuTual \cite{cui2020mutual} & 1 & 8.8K & Single & - & - & $\checkmark$ & $\checkmark$ & $\times$ \\
XPersona ~\cite{lin2021xpersona} & 6 & 104.6K  & Multiple & - & 1,155 & $\checkmark$ & $\times$ & $\times$ \\ 
GlobalWOZ \cite{ding2022globalwoz} & 21 & 9.4K  & Single & - & - & $\checkmark$ & $\times$ & $\times$ \\ 
Multi2WOZ \cite{hung2022multi2woz} & 5 & 1K  & Single & - & - & $\checkmark$ & $\times$ & $\times$ \\
Multi3WOZ \cite{hu2023multi} & 4 & 8.2K  & Single & - & - & $\checkmark$ & $\checkmark$ & $\times$ \\
XDailyDialog \cite{liu2023xdailydialog} & 4 & 52K & Single & 10 & 0 & $\checkmark$ & $\times$ & $\times$ \\ 
\midrule
\textbf{$\datasetname$} & \textbf{8} & \textbf{32K} & \textbf{Multiple} & \textbf{300} & \textbf{210} & \textbf{$\checkmark$} & \textbf{$\checkmark$} & \textbf{$\checkmark$} \\ 
\bottomrule
\end{tabular}
}
\caption{Comparison of dialogue dataset statistics. Our dataset is the only dataset in the list that focuses on generating culture-related entities in the multi-turn conversational dataset. \#Dial. represents the number of dialogues, and $\neg$Translation resembles whether the dataset is from the translation of an existing dataset.}
\label{tab:dialogue_comparison}
\end{table*}

\section{SEADialogues}

$\datasetname$ is a multi-turn, multilingual dialogue dataset encompassing eight Southeast Asian languages (Indonesian, Javanese, Malay, Minangkabau, Tagalog, Tamil, Thai, Vietnamese) from six different countries. For comparison, Table~\ref{tab:dialogue_comparison} highlights how our dataset differs from existing dialogue datasets, with ours being the first to explicitly represent cultural aspects within each conversation.

Figure~\ref{fig:pipeline} illustrates the full data construction pipeline. The process begins with the collection of supporting resources, including scenario and persona templates, along with culturally relevant Southeast Asian names. For each dialogue, two domain-relevant scenarios and corresponding personas are selected to ensure consistency and coherence across both intra-scenario and inter-scenario–persona relationships. In the next step, lexicalization is performed by manually curating and inserting matched entities into both scenarios and personas. This step ensures that subtopics are contextually aligned with the personas and maintain linguistic coherence. Once the templates are finalized, dialogue generation is carried out by prompting LLMs with carefully crafted instructions that incorporate the lexicalized scenarios, personas, and the selected cultural names.

The generated dialogues are subsequently evaluated through a two-fold approach: human annotation for qualitative assessment and automatic evaluation using G-Eval \cite{liu2023g}, M-Prometheus \cite{pombal2025mprometheussuiteopenmultilingual}, and R3 \cite{anugraha2025r3robustrubricagnosticreward} to provide quantitative insights. This structured pipeline ensures the generation of high-quality, culturally appropriate dialogue data for our main objective in this study.

\subsection{Template Generation}

To craft rich, multi-turn dialogue data, we begin by curating several reusable resources we call templates. Mainly, there are two templates to construct the dialogue setup, which are scenario and speakers' persona templates. Both are derived from our curated topics list, which consists of 100 topics in total. To build them, we employ the GPT-4.1 mini~\citep{achiam2023gpt} LLM model to generate 300 scenarios, where 1 topic corresponds to 3 scenarios. For each scenario, we identify culturally grounded entities and replace them with abstract placeholders. These placeholder slots will later be instantiated with values drawn from a curated pool of culturally relevant entities. This approach allows the same dialogue scenario to be adapted across multiple country contexts while maintaining cultural fidelity. After all scenario templates have been generated, we conduct human annotation to identify and revise low-quality or inappropriate templates. See Table~\ref{tab:topic-templates} for the scenario template examples. 

In parallel, we also develop persona templates to characterize the personality traits of each speaker. These templates guide the linguistic expression and behavioral tendencies throughout the dialogue. Similar to topic templates, any culture-specific references within the persona descriptions are masked and later filled with values from a corresponding cultural entity pool. Initially, we aimed to have the same number of personas as scenario templates, which is 300 in total. However, we decided to drop 90 personas due to the low quality of their associated generated templates. See Table~\ref{tab:persona-templates} for the persona template examples.

Each dialogue setup consists of two scenarios. This multi-scenario design mirrors the dynamic nature of real-world conversations, which often shift fluidly between different topics rather than remaining fixed on a single subject. To address this, at the end of this step, we select two scenario templates and two persona templates (per participant in the two-way exchange) for use in the subsequent lexicalization stage for each generated dialogue.




\subsection{Lexicalization}

Lexicalization is essential for embedding cultural nuances into dialogue. To achieve this, 
we incorporate entity lists tagged with language codes, such as \texttt{-ind} for Indonesian, \texttt{-tha} for Thai, or the generic (entities that can be used in all languages) \texttt{-gen}, to indicate their intended scope of use. Examples of these tagged entities are provided in Table~\ref{tab:entities}. Those entities are used to generating specific scenarios by lexicalizing the scenario templates. Using the selected templates from the previous step, each slot is systematically filled with entities whose language tags match the target language of the dialogue. To ensure comprehensive coverage, all valid combinations generated through this slot-filling procedure are enumerated.
For example, to generate Indonesian dialogue, the entity `iconic ricepaddies of Ubud-ind' can be inserted into the template `Person A describes a family trip to the [TRAVEL\_DESTINATION]' to produce: ``Person A describes a family trip to the iconic rice paddies of Ubud.'' Concurrently, speaker personas are generated to populate the dialogues. The slots within this template are then filled with appropriate entities from the compiled lists. To add depth and individuality to each persona, one personality trait is randomly selected from the list of personality traits and appended to the description, thereby enriching the character profile.

\paragraph{Preserve and validate contextual dependencies.}
To ensure the dialogues remain authentic and culturally relevant, contextual alignment between certain slots within a template must be retained. We build a dictionary in JSON files for pairs of slots that are inherently linked either culturally or semantically. It helps exclude combinations that would be incompatible or nonsensical, thereby maintaining the coherence and realism of the dialogue. For instance, if there are both \texttt{[COUNTRY]} and \texttt{[CITY]} entities within a template, for \texttt{[CITY]} values like Jakarta, Bandung, and Denpasar will only be paired with Indonesia as \texttt{[COUNTRY]} value, or Bangkok, Chiang Mai, and Songkhla will only be paired with Thailand.

\subsection{Dialogue Generation}

Once all necessary components are collected, including the target language for the dialogue, lexicalized scenarios, lexicalized personas, personalities, and character names, we proceed to the dialogue generation phase. In this phase, we curate a prompt designed to guide the model in generating culturally appropriate dialogue and input it into both open-source and proprietary LLMs. The prompt used for dialogue generation is shown in Appendix~\ref{sec:topics_personas_appendix}. Additionally, we specify a maximum number of dialogue turns within the prompt to prevent overly long or unnatural conversations.


\subsection{Annotations}
To ensure the quality and validity of the dataset, we employ a structured human annotation process. Each dialogue is assessed based on the following criteria to evaluate conversational abilities: (1) \textit{Fluency}, (2) \textit{Engagingness}, (3) \textit{Coherence}, (4) \textit{Naturalness}, and (5) \textit{Culturally Relevance}. In addition, annotators conduct these evaluations to measure instruction-following abilities: (1) \textit{Profile Detection} and (2) \textit{Correctness}. Detailed annotation scoring rubrics and guidelines are provided in Appendices \ref{sec:annotation_rubrics_appendix} and \ref{sec:annotation_guidelines_appendix}. For each language, three annotators are employed to perform annotations on our platform. They were hired through our contacts and are indigenous people from the country, fluent in the native language. To see an overview of our platform, please refer to Appendix \ref{sec:annotation_platform_appendix}.

\subsection{Automatic Evaluation}
Relying solely on human evaluation poses several challenges, including significant time requirements, logistical complexities, and inherent subjectivity. While human evaluation is our primary measure of dialogue quality, we strategically incorporate automatic evaluation methods. The main goal of using these automatic evaluation methods is to ensure the scalability of our data creation process. 

We utilize G-EVAL \cite{liu2023g}, a prompt-based evaluator as the LLM-as-judge. The prompt provides information regarding the definition of the evaluation task and the assessment criteria. It also employs a chain of thought, consisting of a series of instructions that outline the evaluation steps. Additionally, it includes a scoring function that interacts with a language model using a form-filling approach and probabilities of the return tokens. In addition to G-EVAL, we also use the R3 \cite{anugraha2025r3robustrubricagnosticreward} and M-Prometheus \cite{pombal2025mprometheussuiteopenmultilingual} reward model as the LLM-as-judge. These methods use point-wise evaluation, which assesses the quality of a single response by assigning an integer score based on specific criteria. The reward model takes task instructions, responses, and rubrics as input, generates the scores, and provides explanations with its reasoning model capability. For the details of instructions, rubrics, and prompts, see section \ref{sec:automatic-evaluation-detail} in the Appendix.

\begin{table*}[!ht]
    \centering
    \resizebox{0.95\textwidth}{!}{
    \begin{tabular}{l|c c c c c c c c}
    \toprule
    \textbf{Model} & \texttt{ind} & \texttt{jav} & \texttt{min} & \texttt{tam} & \texttt{tgl} & \texttt{tha} & \texttt{vie} & \texttt{zsm} \\
    \midrule
    \textbf{Coherence} & & & & \\
    \quad Aya-8B-Expanse & \textbf{3.00 ± 0.00} & \textbf{3.00 ± 0.00} & 3.00 ± 0.01 & 2.42 ± 0.55 & 2.34 ± 0.62 & 2.41 ± 0.50 & \textbf{3.00 ± 0.00} & \textbf{3.00 ± 0.00} \\
    \quad Gemini 1.5 Flash & \textbf{3.00 ± 0.00 }& \textbf{3.00 ± 0.00} & \textbf{3.00 ± 0.00} & \textbf{3.00 ± 0.00} & \textbf{3.00 ± 0.00} & \textbf{3.00 ± 0.00} & \textbf{3.00 ± 0.00} & 3.00 ± 0.01 \\
    \quad GPT-4o mini & \textbf{3.00 ± 0.00} & \textbf{3.00 ± 0.00} & \textbf{3.00 ± 0.00} & \textbf{3.00 ± 0.00} & \textbf{3.00 ± 0.00} & \textbf{3.00 ± 0.00} & \textbf{3.00 ± 0.00 }& \textbf{3.00 ± 0.00} \\
    \quad Llama-3.1-Instruct & 3.00 ± 0.01 & 1.97 ± 0.76 & 2.59 ± 0.66 & 2.58 ± 0.44 & 2.57 ± 0.60 & 2.90 ± 0.32 & 2.98 ± 0.10 & 2.99 ± 0.06 \\
    \midrule
    \textbf{Culturally Relevance} & & & & \\
    \quad Aya-8B-Expanse & 2.98 ± 0.10 & 2.98 ± 0.15 & \textbf{2.99 ± 0.04} & 2.55 ± 2.67 & 2.92 ± 5.49 & 1.33 ± 0.81 & 3.00 ± 0.01 & 2.98 ± 0.10 \\
    \quad Gemini 1.5 Flash & \textbf{3.00 ± 0.03} & \textbf{3.00 ± 0.00} & 2.99 ± 0.08 & \textbf{3.00 ± 0.01} & \textbf{3.00 ± 0.02} & \textbf{2.99 ± 0.04} & 3.00 ± 0.01 & \textbf{3.00 ± 0.03} \\
    \quad GPT-4o mini & 2.99 ± 0.03 & 3.00 ± 0.03 & 2.97 ± 0.10 & 3.00 ± 0.02 & 2.99 ± 0.03 & 2.99 ± 0.06 & \textbf{3.00 ± 0.00} & 2.99 ± 0.06 \\
    \quad Llama-3.1-Instruct & 2.95 ± 0.20 & 2.09 ± 2.10 & 2.49 ± 0.82 & 2.25 ± 0.93 & 2.66 ± 0.53 & 2.79 ± 0.42 & 2.92 ± 0.23 & 2.89 ± 0.33 \\
    \midrule
    \textbf{Engagingness} & & & & \\
    \quad Aya-8B-Expanse & \textbf{2.59 ± 0.36} & \textbf{2.62 ± 0.34} & \textbf{2.64 ± 0.34} & 1.70 ± 0.39 & 1.71 ± 0.39 & 1.55 ± 0.32 & \textbf{2.66 ± 0.36} & \textbf{2.56 ± 0.37} \\
    \quad Gemini 1.5 Flash & 2.30 ± 0.35 & 2.31 ± 0.38 & 2.21 ± 0.36 & 2.04 ± 0.43 & 2.08 ± 0.33 & 2.42 ± 0.43 & 2.38 ± 0.38 & 2.26 ± 0.39 \\
    \quad GPT-4o mini & 2.42 ± 0.35 & 2.44 ± 0.35 & 2.35 ± 0.36 & \textbf{2.14 ± 0.31} & \textbf{2.57 ± 0.34} & \textbf{2.45 ± 0.37} & 2.47 ± 0.39 & 2.38 ± 0.38 \\
    \quad Llama-3.1-Instruct & 1.86 ± 0.26 & 1.17 ± 0.33 & 1.52 ± 0.45 & 1.26 ± 0.24 & 1.32 ± 0.34 & 1.58 ± 0.41 & 1.82 ± 0.39 & 1.84 ± 0.33 \\
    \midrule
    \textbf{Fluency} & & & & \\
    \quad Aya-8B-Expanse & 3.00 ± 0.00 & \textbf{3.00 ± 0.00} & \textbf{3.00 ± 0.00} & 2.29 ± 0.50 & 1.45 ± 0.49 & 1.56 ± 0.48 & \textbf{3.00 ± 0.00} & \textbf{3.00 ± 0.00} \\
    \quad Gemini 1.5 Flash & 3.00 ± 0.01 & \textbf{3.00 ± 0.00} & 2.99 ± 0.02 & \textbf{3.00 ± 0.00} & \textbf{3.00 ± 0.00} & \textbf{3.00 ± 0.04} & \textbf{3.00 ± 0.00} & \textbf{3.00 ± 0.00} \\
    \quad GPT-4o mini & 3.00 ± 0.00 & \textbf{3.00 ± 0.00} & \textbf{3.00 ± 0.00} & 2.99 ± 0.06 & \textbf{3.00 ± 0.00} & 2.99 ± 0.09 & \textbf{3.00 ± 0.00} & \textbf{3.00 ± 0.00} \\
    \quad Llama-3.1-Instruct & 3.00 ± 0.01 & 1.60 ± 0.71 & 2.43 ± 0.75 & 2.54 ± 0.44 & 2.31 ± 0.66 & 2.71 ± 0.51 & 2.92 ± 0.25 & 2.97 ± 0.15 \\
    \midrule
    \textbf{Naturalness} & & & & \\
    \quad Aya-8B-Expanse & 3.00 ± 0.03 & \textbf{3.00 ± 0.00} & \textbf{3.00 ± 0.00} & 1.98 ± 0.45 & 1.32 ± 0.35 & 1.21 ± 0.25 & \textbf{3.00 ± 0.00} & \textbf{3.00 ± 0.00} \\
    \quad Gemini 1.5 Flash & 2.98 ± 0.10 & 2.97 ± 0.08 & 2.99 ± 0.05 & 2.97 ± 0.12 & 2.91 ± 0.20 & \textbf{2.98 ± 0.08} & 2.99 ± 0.04 & 2.98 ± 0.07 \\
    \quad GPT-4o mini & \textbf{3.00 ± 0.01} & 2.98 ± 0.12 & 3.00 ± 0.02 & \textbf{2.97 ± 0.10} & \textbf{3.00 ± 0.02} & \textbf{2.98 ± 0.08} & \textbf{3.00 ± 0.00} & 3.00 ± 0.01 \\
    \quad Llama-3.1-Instruct & 2.60 ± 0.46 & 1.17 ± 0.43 & 1.66 ± 0.74 & 1.69 ± 0.44 & 1.43 ± 0.52 & 1.77 ± 0.56 & 2.23 ± 0.69 & 2.33 ± 0.56 \\
    \midrule
    \textbf{Correctness} & & & & \\
     \quad Aya-8B-Expanse & 0.98 ± 0.10 & 0.99 ± 0.04 & 1.00 ± 0.02 & 0.44 ± 0.38 & 0.66 ± 0.34 & 0.48 ± 0.39 & \textbf{1.00 ± 0.00} & 0.98 ± 0.10 \\
    \quad Gemini 1.5 Flash & \textbf{1.00 ± 0.00} & \textbf{1.00 ± 0.01} & \textbf{1.00 ± 0.00} & \textbf{1.00 ± 0.00} & 1.00 ± 0.01 & \textbf{1.00 ± 0.00} & 1.00 ± 0.01 & \textbf{1.00 ± 0.00} \\
    \quad GPT-4o mini & 1.00 ± 0.01 & 1.00 ± 0.02 & 1.00 ± 0.02 & 1.00 ± 0.01 & \textbf{1.00 ± 0.00} & \textbf{1.00 ± 0.00} & 0.99 ± 0.06 & \textbf{1.00 ± 0.00} \\
    \quad Llama-3.1-Instruct & 0.98 ± 0.11 & 0.77 ± 0.35 & 0.91 ± 0.26 & 0.41 ± 0.36 & 0.96 ± 0.13 & 0.97 ± 0.13 & 0.99 ± 0.06 & 0.98 ± 0.09 \\
    \midrule
    \textbf{Profile Detection} & & & & \\
    \quad Aya-8B-Expanse & 0.96 ± 0.13 & 0.98 ± 0.06 & 0.97 ± 0.09 & 0.70 ± 0.27 & 0.63 ± 0.30 & 0.51 ± 0.27 & \textbf{0.99 ± 0.06} & 0.96 ± 0.11 \\
    \quad Gemini 1.5 Flash & \textbf{0.97 ± 0.10} & 0.98 ± 0.06 & \textbf{0.99 ± 0.04} & \textbf{0.99 ± 0.04} & 0.95 ± 0.13 & 0.98 ± 0.06 & 0.98 ± 0.08 & 0.95 ± 0.14 \\
    \quad GPT-4o mini & 0.96 ± 0.14 & \textbf{0.99 ± 0.03} & 0.96 ± 0.12 & 0.97 ± 0.09 & 0.97 ± 0.09 & 0.96 ± 0.12 & 0.98 ± 0.08 & 0.98 ± 0.07 \\
    \quad Llama-3.1-Instruct & 0.93 ± 0.17 & 0.59 ± 0.35 & 0.80 ± 0.29 & 0.70 ± 0.28 & 0.78 ± 0.25 & 0.83 ± 0.23 & \textbf{0.97 ± 0.07} & 0.94 ± 0.11 \\
    \bottomrule
    \end{tabular}
    }
    \caption{G-Eval results on conversational generation and instruction following capabilities. Coherence, Engagingness, Fluency, and Naturalness have a score range of 1 to 3, while Culturally Relevance has a score range of 0 to 3. For instruction-following metrics, Profile Detection and Correctness have a binary score.}
    \label{tab:automatic-metrics}
\end{table*}

\section{Experimental Setup}

\subsection{Dialogue Generation}
We employ four different models for dialogue generation, which include both closed-source and open-source options. For our open-source models, we use Llama-3.1-8B Instruct \cite{grattafiori2024llama} and Aya-8B Expanse \cite{li2017dailydialog}. For the closed-source models, we utilize Gemini-Flash-1.5 \cite{team2024gemini} and GPT-4o mini \cite{achiam2023gpt}. We run these models on an Nvidia A100 40GB and use HuggingFace, OpenAI, and Google API as packages to run these models.
In line with best practices observed in prior work on dialogue generation using LLMs ~\citep{li-2024-dynamic, ye2024lstdial}, we search for the best hyperparameter by comparing the dialogue, and set the sampling parameters to a temperature (\textit{T}) of 0.7 and a (\textit{top-P}) of 0.8. At the end of the pipeline, we successfully collected a total of 32,000 dialogues, distributed evenly as 4,000 dialogues per language, with each model contributing a unified set of 1,000 dialogues under this generation scheme.

\subsection{Automatic Evaluation}
We utilize automatic evaluation through LLM-as-judge methods, including G-Eval \cite{liu2023g} as our primary method, M-Prometheus \cite{pombal2025mprometheussuiteopenmultilingual}, and R3 reward model \cite{anugraha2025r3robustrubricagnosticreward} in our dataset. For G-Eval, we use the GPT-4.1 mini model; for M-Prometheus, we apply the M-Prometheus-7B model; and for R3 reward model, we utilize the R3-Qwen-14B-14k version. These automatic methods assess the generated dialogues using the same metrics that we applied during our human annotation process.

\begin{table}[!th]
    \centering
    \resizebox{0.49\textwidth}{!}{
    \begin{tabular}{l|cccccccc}
    \toprule
    \textbf{Model} & \texttt{ind} & \texttt{jav} & \texttt{min} & \texttt{tha} \\
    \midrule
    \textbf{Coherence} \\
    \quad Aya-8B-Expanse & 2.79 ± 0.44 & 2.84 ± 0.43 & 2.81 ± 0.40 & 1.98 ± 0.73 \\
    \quad Gemini 1.5 Flash   & \textbf{2.88 ± 0.34} & 2.90 ± 0.34 & \textbf{2.82 ± 0.40} & \textbf{2.46 ± 0.70} \\
    \quad GPT-4o mini        & 2.84 ± 0.37 & \textbf{2.92 ± 0.30} & 2.75 ± 0.45 & 2.45 ± 0.64 \\
    \quad Llama-3.1-Instruct & 2.70 ± 0.49 & 1.17 ± 0.48 & 2.41 ± 0.73 & 2.19 ± 0.75 \\
    \midrule
    \textbf{Culturally Relevance} \\
    \quad Aya-8B-Expanse     & 2.02 ± 0.98 & 1.93 ± 1.29 & 2.15 ± 1.04 & 0.71 ± 0.93 \\
    \quad Gemini 1.5 Flash   & \textbf{2.30 ± 0.86} & 2.22 ± 1.19 & \textbf{2.17 ± 0.98} & \textbf{1.59 ± 1.15} \\
    \quad GPT-4o mini        & 2.11 ± 0.92 & \textbf{2.23 ± 1.18} & 2.16 ± 1.04 & 1.42 ± 1.15 \\
    \quad Llama-3.1-Instruct & 1.95 ± 0.88 & 0.36 ± 0.80 & 1.88 ± 1.16 & 1.37 ± 1.12 \\
    \midrule
    \textbf{Engagingness} \\
    \quad Aya-8B-Expanse & \textbf{2.62 ± 0.52} & \textbf{2.75 ± 0.47} & \textbf{2.72 ± 0.45} & 1.61 ± 0.59 \\
    \quad Gemini 1.5 Flash & 2.51 ± 0.52 & 2.56 ± 0.54 & 2.62 ± 0.49 & \textbf{2.26 ± 0.70} \\
    \quad GPT-4o mini & 2.35 ± 0.50 & 2.68 ± 0.49 & 2.58 ± 0.52 & 2.04 ± 0.72 \\
    \quad Llama-3.1-Instruct & 2.10 ± 0.63 & 1.21 ± 0.55 & 2.30 ± 0.71 & 1.93 ± 0.60 \\
    \midrule
    \textbf{Fluency} \\
    \quad Aya-8B-Expanse  & 2.75 ± 0.47  & 1.09 ± 0.41 & 1.48 ± 0.73 & 1.70 ± 0.58 \\
    \quad Gemini 1.5 Flash & \textbf{2.92 ± 0.28} & 2.72 ± 0.48 & \textbf{2.32 ± 0.54} & \textbf{2.30 ± 0.60} \\
    \quad GPT-4o mini & 2.88 ± 0.32 & \textbf{2.73 ± 0.47} & 1.46 ± 0.68 & 2.15 ± 0.61 \\
    \quad Llama-3.1-Instruct & 2.83 ± 0.39 & 1.04 ± 0.22 & 1.50 ± 0.59 & 1.98 ± 0.57 \\
    \midrule
    \textbf{Naturalness} \\
    \quad Aya-8B-Expanse     & 2.49 ± 0.57 & 1.46 ± 0.69 & 2.34 ± 0.52 & 1.35 ± 0.48 \\
    \quad Gemini 1.5 Flash   & \textbf{2.60 ± 0.52} & \textbf{2.30 ± 0.76} & \textbf{2.46 ± 0.53} & \textbf{2.15 ± 0.53} \\
    \quad GPT-4o mini        & 2.28 ± 0.47 & 2.26 ± 0.75 & 2.24 ± 0.55 & 1.83 ± 0.50 \\
    \quad Llama-3.1-Instruct & 2.09 ± 0.48 & 1.02 ± 0.14 & 1.91 ± 0.72 & 1.64 ± 0.48 \\
    \midrule
    \textbf{Correctness} \\
    \quad Aya-8B-Expanse & 0.91 ± 0.28 & 0.97 ± 0.17 & 0.99 ± 0.08 & 0.65 ± 0.48 \\
    \quad Gemini 1.5 Flash   & \textbf{0.98 ± 0.15} & 0.96 ± 0.19 & \textbf{1.00 ± 0.06} & 0.88 ± 0.32 \\
    \quad GPT-4o mini        & \textbf{0.98 ± 0.15} & \textbf{0.98 ± 0.13} & 0.99 ± 0.08 & \textbf{0.91 ± 0.29} \\
    \quad Llama-3.1-Instruct & 0.90 ± 0.30 & 0.25 ± 0.43 & 0.86 ± 0.34 & 0.87 ± 0.33 \\
    \midrule
    \textbf{Profile Detection} \\
    \quad Aya-8B-Expanse     & \textbf{0.65 ± 0.48} & 0.81 ± 0.39 & \textbf{0.95 ± 0.22} & 0.40 ± 0.49 \\
    \quad Gemini 1.5 Flash   & 0.57 ± 0.49 & 0.80 ± 0.40 & 0.92 ± 0.27 & \textbf{0.81 ± 0.39} \\
    \quad GPT-4o mini        & 0.59 ± 0.49 & \textbf{0.83 ± 0.38} & 0.94 ± 0.25 & 0.77 ± 0.42 \\
    \quad Llama-3.1-Instruct & 0.59 ± 0.49 & 0.19 ± 0.39 & 0.81 ± 0.40 & 0.71 ± 0.45 \\
    \bottomrule
    \end{tabular}
    }
    \caption{Human Annotations Results.}
    \label{tab:human-annotations}
\end{table}

\section{Results and Analysis}

\subsection{Human Evaluation}
As shown in Table~\ref{tab:human-annotations}, in terms of conversational capabilities, both GPT-4o Mini and Gemini 1.5 Flash demonstrate the best performance across various metrics. However, Gemini 1.5 Flash leads on most metrics, particularly in fluency for the Minangkabau and Thai languages. Among the open-source models, Aya-8B-Expanse performs well with Javanese and Minangkabau, though it falls short on the fluency metric. In contrast, Llama-3.1 Instruct generally produces the lowest scores across most metrics.

For the following instructions abilities, Gemini 1.5 Flash and GPT-4o mini exhibit good performance, nearly 100 percent accuracy on Correctness. However, they have lower performance on Profile Detection metrics. For Aya-8B-Expanse, it shows similar performance with Gemini 1.5 Flash and GPT-4o mini for Indonesian, Javanese, and Minangkabau, but struggles with Thai. Lastly, in the case of Llama-3.1 Instruct, its overall performance is generally lower, except when compared to Aya-8B-Expanse in Thai.

\begin{figure*}[!th]
    \centering
    \includegraphics[width=1\linewidth]{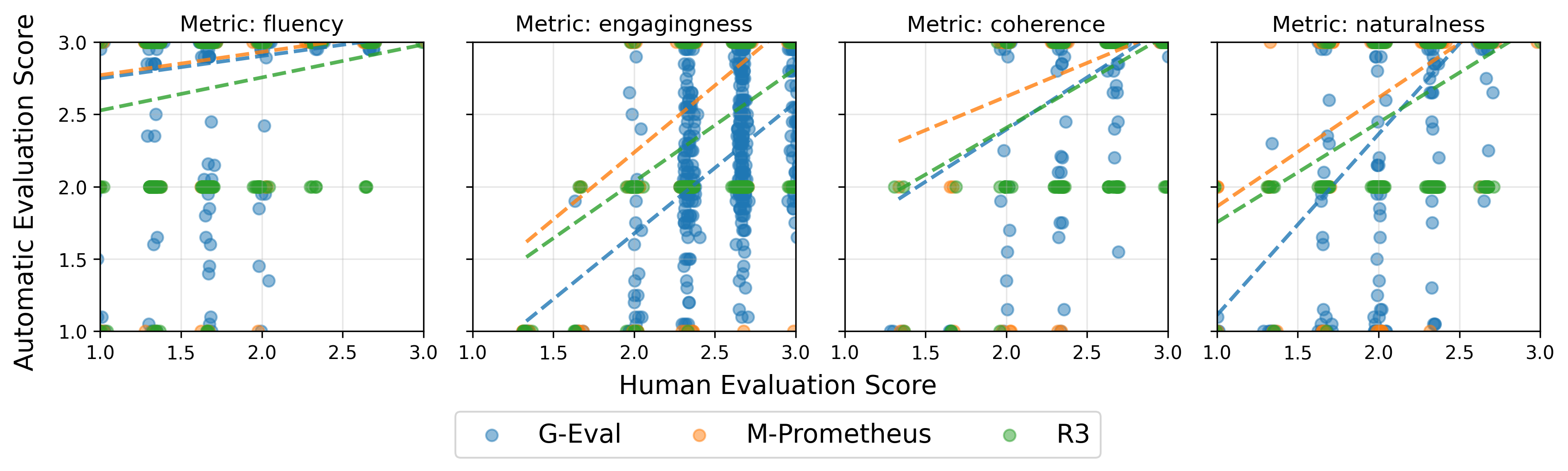}
    \caption{Correlation between Automatic Evaluations and Human Annotations per Metric on Minangkabau.}
    \label{fig:corr-conversational-quality}
\end{figure*}

\begin{table}[!th]
    \centering
    \resizebox{0.49\textwidth}{!}{
    \begin{tabular}{l|c c c}
    \toprule
    \textbf{} & \textbf{G-Eval} & \textbf{M-Prometheus} & \textbf{R3} \\
    
    \midrule
    \textbf{Pearson} \\
    \quad Coherence & \textbf{0.5876} & 0.4734 & 0.5227  \\
    \quad Culturally Relevance & \textbf{0.4832} & 0.4324 & 0.3027 \\
    \quad Engagingness & 0.5401 & \textbf{0.5499} & 0.4627 \\
    \quad Fluency & 0.1528 & 0.1749 & \textbf{0.1949} \\
    \quad Naturalness & \textbf{0.6124} & 0.5228 & 0.4501 \\
    \midrule
    \textbf{Spearman} \\
    \quad Coherence & \textbf{0.4408} & 0.3088 & 0.3994  \\
    \quad Culturally Relevance & \textbf{0.2506} & 0.2241 & 0.2146 \\
    \quad Engagingness & \textbf{0.4486} & 0.4462 & 0.3754 \\
    \quad Fluency & 0.0372 & 0.1346 & \textbf{0.1429} \\
    \quad Naturalness & \textbf{0.5307} & 0.4650 & 0.3830 \\
    \midrule
    \textbf{Kendall Tau} \\
    \quad Coherence & \textbf{0.4100} & 0.2893 & 0.3765  \\
    \quad Culturally Relevance & \textbf{0.2234} & 0.2069 & 0.1969 \\
    \quad Engagingness & 0.3588 & \textbf{0.4098} & 0.3482 \\
    \quad Fluency & 0.0326 & 0.1220 & \textbf{0.1300} \\
    \quad Naturalness & \textbf{0.4660} & 0.4270 & 0.3551 \\
    \bottomrule
    \end{tabular}
    }
    \caption{Automatic Evaluations and Human Annotations Correlations on Minangkabau.}
    \label{tab:auto_human_correlations}
\end{table}

\subsection{Automatic Evaluation}
In terms of conversational quality, as shown in Table \ref{tab:automatic-metrics}, closed-weight models like Gemini 1.5 Flash and GPT-4o mini generally achieve higher scores across most metrics and languages. In contrast, open-weight models, particularly Llama-3.1-Instruct, display more variability and show lower performance. On a positive note, the Aya-8B-Expanse model receives comparably high scores for languages such as Indonesian, Javanese, Minangkabau, Vietnamese, and Malay, ranking among the best for engagingness and naturalness in these languages.

For instruction-following to generate the dialogue capabilities, Gemini 1.5 Flash and GPT-4o mini demonstrate strong results in both correctness and profile detection. Llama-3.1-Instruct performs adequately in correctness for specific languages, although it struggles with Tamil and Javanese. However, it shows poorer results in profile detection for most languages. Similar to its conversational capabilities, Aya-8B-Expense shows promising results in both correctness and profile detection for Indonesian, Javanese, Minangkabau, Vietnamese, and Malay, but performs less effectively with other languages. These results indicate similar trends in both Human Evaluation and Automatic Evaluation, particularly in pointing out which model can generate good dialogue for specific languages. See the result details in Appendix~\ref{sec:automatic-evaluation-detail}.

\subsection{Human Annotation Alignment}

Figure \ref{fig:corr-conversational-quality} presents a visual analysis of the correlation between the automatic evaluation scores and human judgments for key conversational quality metrics, such as fluency, coherence, engagingness, and naturalness. All metrics show positive correlations between automatic and human evaluations. G-Eval aligns more closely with human judgment than other automatic evaluation methods, as shown in Table \ref{tab:auto_human_correlations}. The most significant challenges lie in the Fluency and Cultural Relevance metrics, as evidenced by their lower correlation compared to other metrics.

\section{Related Work}

\subsection{Personalized Conversations}

Dialogue research has primarily focused on task-oriented systems~\cite{budzianowski2018multiwoz,ding2022globalwoz,goel2023presto,he2024multiif} and question-answering dialogues~\cite{feng2020doc2dial,rajpurkar2016squad}, but these approaches often lack the richness of open-domain, human-human interactions. Datasets like DailyDialog~\cite{li2017dailydialog} and XDailyDialog~\cite{liu2023xdailydialog} address this by offering multi-turn, intent- and emotion-annotated dialogues for more naturalistic chit-chat.

Building on this, persona-conditioned datasets such as PERSONA-CHAT~\cite{zhang2018personalizing} and PersonalDialog~\cite{zheng2019personalized} simulate personalized conversations using user profiles. Recent work like BotChat~\cite{duan2023botchat} uses LLMs to generate scalable, persona-driven dialogues from seed prompts. Multilingual datasets, including XPersona~\cite{lin2021xpersona} and XDailyDialog, improve cross-lingual transfer through translated and refined dialogues. While some datasets constrain dialogue to a single topic, others like PERSONA-CHAT and BotChat allow topic shifts, better mirroring real-world conversation dynamics. Finally, cultural grounding has gained importance. Efforts like GlobalWOZ~\cite{ding2022globalwoz} localize templates to reflect cultural norms—a principle we also adopt in our dataset design.

\subsection{Dataset Evaluation}

Recent work increasingly adopts LLM-based evaluation to complement human assessment. BotChat~\cite{duan2023botchat} introduces a three-part framework: UniEval for single-model judgments, BotChat Arena for pairwise comparisons, and G-Eval~\cite{liu2023g} for aligning model outputs with human references. XDailyDialog~\cite{liu2023xdailydialog} combines automatic metrics (e.g., BLEU, F1, DIST-n) with human evaluations at both turn and dialogue levels, assessing fluency, relevance, and coherence. Inspired by these practices, we adopt a hybrid evaluation protocol that integrates human ratings with LLM-based scoring using G-Eval, enabling scalable and consistent quality assessment.

\section{Conclusion}
We introduce \datasetname, an open-source, multilingual, multi-turn, and persona-rich synthetic dialogue dataset that spans eight languages across six Southeast Asian countries. Motivated by the impressive generative capabilities of large language models, 
we incorporates structured guardrails in the data generation pipeline, including scenario and persona templates as well as culturally grounded lexicalization strategies. We perform further study on samples from our dataset, by conducting a comprehensive evaluation using both human annotation and LLM-as-a-judge assessments across several key metrics. These include fluency, coherence, naturalness, engagingness, cultural relevance, persona consistency, and factual correctness. Our findings indicate that proprietary (closed-weight) models generally outperform open-weight models on these dimensions. This highlights a pressing need for high-quality, culturally enriched, and persona-aware datasets like ours to support the development of open-weight LLMs. In turn, such resources can help bridge the quality gap with proprietary models. Additionally, improving the cultural and persona grounding in datasets may enhance the alignment of LLM-as-a-judge systems with human evaluators in future dialogue assessment tasks.

\section*{Acknowledgments}
We thank all annotators whose effort and expertise were essential to the construction of our dataset. We also would like to thank Chulalongkorn University's AI Center (Chula.AI) on the annotation support for Thai.

\section*{Limitations}
This paper focuses on approximating natural human-human conversations using large language models. However, it does not yet include targeted evaluations on specific benchmarks such as topic transition detection \cite{soni2022empirical} or persona detection \cite{jun2025exploring}. These tasks are particularly relevant for capturing conversational nuances in Southeast Asian cultural contexts. While this work represents an initial step toward understanding LLM-generated dialogue in Southeast Asian settings, future work should incorporate more comprehensive benchmarking and explore task-specific methodologies to better assess and improve performance in culturally grounded dialogue generation.

Additionally, although humans manually curate the topics, names, and entities, there is still a risk that the generated dialogues may not fully capture the cultural nuances of certain languages, as they are produced using various LLMs. However, we make every effort to ensure cultural appropriateness by carefully curating the seed entities and related content for each region.

\section*{Ethics Statement}
Throughout our study, we commit to adhering to ethical standards and best practices in NLP research. The dialogue data includes character names selected from manually curated name lists representative of each country. These name pools are created using publicly available, non-sensitive sources (e.g., common baby name registries), and do not reference or target any real individuals. While care has been taken to ensure these are generic, there remains a small possibility that some names may coincidentally match real individuals; any such resemblance is purely coincidental.

To minimize harm, all dialogue topics were manually curated to avoid discussions involving violence or other sensitive content. As a result, we believe the likelihood of harmful or inappropriate material appearing in the dataset is very low. All annotators were compensated fairly, following wage standards in their respective countries. Additionally, all annotators agree for their annotations to be publicly released in an aggregated form (e.g., scores), with no personally identifiable information included. We have taken steps to ensure that annotator privacy and confidentiality are fully maintained.


\bibliography{custom}

\appendix

\section{Dataset Statistics}
\label{sec:dataset_statistics}
Table \ref{tab:seadialogue-statistics} shows the detailed statistics of \datasetname. 
\begin{table}[htbp]
    \centering
    \begin{tabular}{lc}
    \toprule
    No. of languages                 & 8         \\ 
    No. of dialogues                 & 32,000    \\ 
    Average dialogues per language   & 4,000     \\ 
    Average utterances per dialogue  & 13.86     \\ 
    Average words per utterance      & 21.69     \\ 
    No. of topics                    & 100       \\ 
    Topics per dialogue              & 2         \\
    No. of scenarios                 & 300       \\
    No. of personas                  & 210       \\ 
    \bottomrule
    \end{tabular}
    \caption{SEADialogue statistics.}
    \label{tab:seadialogue-statistics}
\end{table}

\section{Persona and Topics}
\label{sec:topics_personas_appendix}
This section presents examples of documents used for data generation within our framework.
\begin{itemize}
    \item Table \ref{tab:topic-templates} presents samples of scenario templates for dialogue.
    \item Table \ref{tab:persona-templates} provides examples of persona templates.
    \item Table \ref{tab:entities} lists entities use to lexicalize the scenario and persona templates.
    \item Table \ref{tab:personality} showcases various personality traits to complete the persona information.
    \item Table \ref{tab:list-of-names-long} contains samples of names for the characters in the dialogue.
    \item Table \ref{tab:ceremony-food-mapping} shows the mapping of coupled entities, specifically when a scenario or persona template includes coupled delexicalized entities.
    \item Finally, Table \ref{tab:prompt} displays the prompt utilized for generating dialogue. 
\end{itemize}

\begin{table*}[p]
\centering
\small
\renewcommand{\arraystretch}{1.2} 
\setlength{\tabcolsep}{4pt}       
\begin{tabularx}{\textwidth}{@{} p{0.18\textwidth} >{\raggedright\arraybackslash}X >{\raggedright\arraybackslash}X >{\raggedright\arraybackslash}X @{}}
\toprule
\textbf{Topic} & \textbf{Scenario Template Example 1} & \textbf{Scenario Template Example 2} & \textbf{Scenario Template Example 3} \\
\midrule
Favorite TV Shows from Childhood & Two people discuss the influence of [LANGUAGE] folklore in their favorite childhood TV shows. & Person A loved a popular [LANGUAGE] [TV\_SHOWS-1], while Person B grew up watching [LANGUAGE] [TV\_SHOWS-2] on TV. & Both discuss how [LANGUAGE] TV shows shaped their childhood and how modern TV differs from those days. \\
Favorite Musicians or Bands & Both people grew up listening to the same iconic singer, [SINGER]. & Person A admires [SONG\_TYPE-1] music, while Person B prefers the uniqueness of [SONG\_TYPE-2]. & They discuss how traditional [LANGUAGE] songs influenced their favorite [SONG\_TYPE] songs nowadays.\\
Movie or Series Characters That Inspire You & Two people discuss how [LANGUAGE] action films' strong female leads inspired them to be more assertive in life. & Person A admires [LANGUAGE] [MOVIE\_TYPE-1] movie characters, while Person B finds inspiration from modern [LANGUAGE] [MOVIE\_TYPE-2] TV series. & [LANGUAGE] mythology-based movies, and how characters rooted in local legends shaped their personal values.\\
The Most Interesting Local Folk Tales or Myths & Both people share stories about [MYTH\_CHARACTER], the [LANGUAGE] legend myth, but one believes in her protective power while the other sees her as just a legend. & Person A is fascinated by the [LANGUAGE] [MYTH\_CHARACTER-1], while Person B prefers [LANGUAGE] tales of [MYTH\_CHARACTER-2]. & Comparing the morals behind [LANGUAGE] folk tales, focusing on [MYTH\_CHARACTER-1] vs [MYTH\_CHARACTER-2].\\
First Experience Watching a Movie in the Cinema & Two people discussing their shared excitement of watching an action movie in a small-town [LANGUAGE] cinema for the first time. & Person A was terrified by the loud sound system in a [CITY] cinema, while Person B found it thrilling and immersive. & Memorable experiences at classic [CITY] cinema chains and how they shaped their love for movies.\\
\bottomrule
\end{tabularx}
\caption{\centering Example of scenario templates used in the dialogue construction process. Templates include delexicalized placeholders (e.g., [LANGUAGE], [TV\_SHOWS]) for later lexicalization.}
\label{tab:topic-templates}
\end{table*}

\begin{table*}[p]
\centering
\small
\renewcommand{\arraystretch}{1.2} 
\setlength{\tabcolsep}{4pt}       
\begin{tabularx}{\textwidth}{@{} p{0.18\textwidth} >{\raggedright\arraybackslash}X >{\raggedright\arraybackslash}X >{\raggedright\arraybackslash}X @{}}
\toprule
\textbf{Topic} & \textbf{Persona Template Example 1} & \textbf{Persona Template Example 2} & \textbf{Persona Template Example 3} \\
\midrule
Favorite TV Shows from Childhood & A person fascinated by traditional [MOVIE\_TYPE] and mythological characters: [MYTH\_CHARACTER] & A person who loved animated [MOVIE\_TYPE] movie & A person who values [MOVIE\_TYPE] TV shows \\
Favorite Musicians or Bands & A nostalgic [SONG\_TYPE] lover who enjoys live performances & A classically trained musician who is fascinated by folk instruments: [TRADITIONAL\_INSTRUMENT] & A person who enjoys discovering [MUSIC\_GENRE] songs from various culture \\
Movie or Series Characters That Inspire You & An energetic extrovert who loves [MOVIE\_TYPE]-packed movies & A thoughtful introvert who enjoys [MOVIE\_TYPE] & A person who appreciates movie characters inspired by folklore and traditional values \\
The Most Interesting Local Folk Tales or Myths & Enthusiast of historical accuracy who loves researching the real events behind myths. & A skeptic person who enjoys listening to stories of [MYTH\_CHARACTER] & A passionate storyteller who interested in myth \\
First Experience Watching a Movie in the Cinema & A person who likes [STATS\_TYPE] movies & An adventurous moviegoer who likes [STATS\_TYPE] theater & A person who likes [ENVIRONMENT\_CONDITION] places \\
\bottomrule
\end{tabularx}
\caption{\centering Example of persona templates used in the dialogue construction process. Templates include delexicalized placeholders (e.g., [LANGUAGE], [TV\_SHOWS]) for later lexicalization.}
\label{tab:persona-templates}
\end{table*}

\begin{table*}[p]
  \small
  \centering
  \resizebox{0.49\textwidth}{!}{
  \begin{tabular}{@{} l l @{}}
    \toprule
    \textbf{Delexicalized} & \textbf{Lexicalized} \\
    \midrule
    \multirow{8}{*}{[LANGUAGE]}
      & Thai-tha \\
      & Indonesian-ind \\
      & Javanese-jav \\
      & Minangkabau-min \\
      & Tagalog-tag \\
      & Malay-mal \\
      & Tamil-tam \\
      & Vietnamese-vie \\
    \hline
    \multirow{24}{*}{[TV\_SHOWS]}
      & drama-gen \\
      & wayang\_(puppet\_show)-ind \\
      & wayang\_(puppet\_show)-jav \\
      & historical\_drama-jav \\
      & folklore\_series-min \\
      & minang\_comedy-min \\
      & cooking\_show-tha \\
      & comedy\_sketch-ind \\
      & mystery\_thriller-gen \\
      & supernatural\_fable-tha \\
      & variety\_show-tag \\
      & sitcom-tag \\
      & musical\_drama-mal \\
      & historical\_fiction-mal \\
      & comedy\_series-mal \\
      & family\_drama-mal \\
      & tamil\_serial-tam \\
      & musical\_program-tam \\
      & talk\_show-tam \\
      & reality\_show-tam \\
      & cai\_luong-vie \\
      & tuong-vie \\
      & cheo-vie \\
      & ho\_chi\_minh\_biopic-vie \\
    \bottomrule
  \end{tabular}}
  \caption{Delexicalized and Lexicalized Entities.}
  \label{tab:entities}
\end{table*}

\begin{table*}[p]
  \small
  \centering
  \resizebox{0.49\textwidth}{!}{
  \begin{tabular}{@{} l l l @{}}
    \toprule
    \multicolumn{3}{c}{\textbf{Personality Traits}} \\
    \midrule
    Active        & Appreciative   & Considerate   \\
    Creative      & Friendly       & Honest        \\
    Imaginative   & Open           & Patient       \\
    Witty         & Ambitious      & Amusing       \\
    Boyish        & Businesslike   & Determined    \\
    \bottomrule
  \end{tabular}}
  \caption{Personality Traits.}
  \label{tab:personality}
\end{table*}

\begin{table*}[p]
    \centering
    \begin{tabular}{cccc}
        \toprule
        \textbf{Gender} & \textbf{Language} & \textbf{First Name} & \textbf{Last Name} \\
        \midrule
        Male & Indonesian & Andi & Setiawan \\
        & & Budi & Hidayat \\
        & & Joko & Saputra \\
        \addlinespace
        Male & Javanese & Agus & Nugraha \\
        & & Mukhti& Wijaya \\
        & & Eko & Wicaksana \\
        \addlinespace
        Male & Minangkabau & Zulkifli & Chaniago \\
        & & Fadli & Rasyid \\
        & & Yusuf & Putra \\
        \addlinespace
        Male & Thai & Ananda & Chaiya \\
        & & Athit & Anuman \\
        & & Chayaphon & Bun Ma \\
        \addlinespace
        Female & Indonesian & Nafisah & Yasmin \\
        & & Dewi & Rahayu \\
        & & Intan & Wahyuni \\
        \addlinespace
        Female & Javanese & Maya & Whidia \\
        & & Gita & Jelita \\
        & & Kartika & Indriani \\
        \addlinespace
        Female & Minangkabau & Nurul & Hasna \\
        & & Laila & Atiqah \\
        & & Citra & Azizah \\
        \addlinespace
        Female & Thai & Atchara & Channarong \\
        & & Kanlaya & Kaew Buasai \\
        & & Kamala & Nunphakdi \\
        \bottomrule
    \end{tabular}
    \caption{Sample List of Names.}
    \label{tab:list-of-names-long}
\end{table*}

\begin{table*}[htbp]
    \centering
    \begin{tabular}{@{}p{5.2cm}p{10.5cm}@{}}
        \toprule
        \textbf{[CEREMONY]} & \textbf{[FOOD]} \\ \midrule
        Hari\_Raya-ind & ketupat-ind; rendang-min; satay-ind \\
        Hari\_Raya-jav & ketupat-ind; gudeg-jav; soto-ind \\
        Hari\_Raya-min & ketupat-ind; rendang-min; dendeng\_batokok-min; ayam\_pop-min \\
        Satu\_Suro-jav & gudeg-jav; nasi\_liwet-jav; tongseng-jav \\
        Loy\_Krathong-tha & pad\_thai-tha; green\_curry-tha; mango\_sticky\_rice-tha \\
        Songkran-tha & som\_tam-tha; tom\_yum-tha \\
        Eid-gen & ketupat-ind; rendang-min; biryani-indic \\
        Sham\_el\_Nessim-eg & ful\_medames-eg; molokhia-eg \\
        Chinese\_New\_Year-chi & pecking\_duck-chi; xialongbao-chi; dim\_sum-chi \\
        Lantern\_Festival-chi & dim\_sum-chi; xialongbao-chi \\
        Diwali-indic & biryani-indic; samosa-indic \\
        Holi-indic & samosa-indic; biryani-indic \\
        Pasko-tag & adobo-tag; lechon-tag \\
        Deepavali-tam & fish\_head\_curry-tam; roti\_prata-tam \\
        Tet-vie & pho-vie; banh\_mi-vie; goi\_cuon-vie \\
        Ramadan\_markets-gen & ketupat-ind; satay-ind; rendang-min \\
        Indonesian\_Independence\_Day-ind & nasi\_goreng-ind; gado\_gado-ind \\
        Turun\_Mandi-min & rendang-min; sate\_padang-min \\
        Kaharian\_ng\_Bagong\_Taon-mal & nasi\_lemak-mal \\
        Banh\_Chung-vie & pho-vie; banh\_mi-vie \\
        Tahun\_Baru\_Cina-mal & nasi\_lemak-mal \\
        Hari\_raya-mal & nasi\_lemak-mal; satay-mal; laksa-mal \\ 
        \bottomrule
    \end{tabular}
    \caption{Mapping between FOOD and CEREMONY entites.}
    \label{tab:ceremony-food-mapping}
\end{table*}

\begin{table*}[p]
  \centering
  \begin{tabular}{l}
    \toprule
        \textbf{Prompt Template} \\
    \midrule
        Create a multi-turn conversation in \{LANGUAGE\} from 2 people where the topic is: \{TOPIC\_1\}, and then move \\ to the topic: \{TOPIC\_2\}. You must only speak in \{LANGUAGE\}. The conversation is in a polite setting. During \\ the conversation, the speaker calls the other with honorifics. \\ \\ Persona Person A (name = \{NAME\_1\}):\\- A \{PERSONALITY\_1\} \{GENDER\_1\}\\- \{persona\_1\}\\ \\Persona Person B (name = \{NAME\_2\}):\\- A \{PERSONALITY\_2\} \{GENDER\_2\}\\- \{PERSONA\_2\}\\ \\Limit the conversation to \{num\_of\_turns\} turns. Please be direct in generating the conversation; do \\ not generate anything except the conversation itself. Because at least there is one topic transition \\ in the conversation, please denote it with a special token [TRANSITION] inside the conversation. \\ Make the transition as smooth as possible.\\ \\For every turn, please follow this format `name: utterance` \\
    \bottomrule
  \end{tabular}
  \caption{Our prompt to generate the dialogues using LLMs.}
  \label{tab:prompt}
\end{table*}

\section{Dialogue Generation Details}
Incorporating multiple scenarios into a single dialogue prompt raises the risk of abrupt or unnatural topic transitions. To address this, we employ \textsc{Top2Vec} \cite{angelov2020top2vec}, a topic modeling method that clusters our curated pool of scenarios based on semantic similarity. This enables us to select pairs of scenarios from within the same cluster, ensuring thematic coherence and smoother transitions between topics.

\section{Annotation Rubrics}
\label{sec:annotation_rubrics_appendix}
To assess the quality of generated dialogues, we employ a comprehensive evaluation rubric (see Table \ref{tab:dialogue_rubrics}) consisting of six criteria: Fluency, Engagingness, Coherence, Naturalness, Cultural Relevance, Profile Detection, and Correctness. Each criterion targets a specific aspect of conversational quality, ensuring both linguistic and contextual alignment with the intended design of the dialogue system.

\section{Human Annotation Guidelines}
\label{sec:annotation_guidelines_appendix}
Each annotation unit consists of a prompt and a multi-turn dialogue generated in response. The prompt includes two topics and two speaker personas. Annotators evaluate the dialogue with respect to quality, persona alignment, and topic relevance through the following steps:

\paragraph{Step 1: Read the Prompt}
Annotators first read the prompt to identify the two intended \textbf{topics} and the two \textbf{personas}, which include details such as speaker background, personality traits, and interests. Cultural or linguistic context (e.g., Indonesian, Javanese, Minangkabau, or Thai) should also be noted when applicable.

\paragraph{Step 2: Read the Dialogue}
Annotators then read the full multi-turn dialogue to understand its tone, structure, and alignment with the prompt. This step ensures that judgments are based on the overall flow and not isolated turns.

\paragraph{Step 3: Score the Dialogue}
Each dialogue is then rated based on the annotation rubrics as mentioned in Appendix \ref{sec:annotation_rubrics_appendix}.

\paragraph{Step 4: Use Examples}
The guideline includes detailed examples in English, Indonesian, Minangkabau, and Thai for each score level, helping annotators apply the criteria consistently across languages and domains.

\begin{table*}[ht]
  \centering
  \small
  \begin{tabular}{@{} l l p{4.0cm} p{8.0cm}@{}}
    \toprule
    \textbf{Criterion} & \textbf{Score} & \textbf{Objective} & \textbf{Evaluation Criteria}  
    \\
    \midrule
    Fluency & 1–3 & Evaluate grammatical correctness and sentence structure. & Review each utterance for grammar, spelling, and structure. Informal language is acceptable if grammatically correct. 
    \newline
    \textbf{- 1: Poor.} The dialogue is poorly constructed, with significant grammar and language issues; it is difficult to understand. \newline
    \textbf{- 2: Fair.} The dialogue is understandable but has some fluency issues. \newline
    \textbf{- 3: Good}. The dialogue is fluent, natural, and easy to read, with minimal or no issues.
    \\
    Engagingness & 1–3 & Assess the depth and interest of the conversation. & Examine the entire dialogue for richness, interaction quality, and ability to sustain interest. 
    \newline
    \textbf{- 1: Poor}. The conversation is flat, boring, and shallow. \newline
    \textbf{- 2: Fair}. Somewhat engaging but the story lacks challenge or flat. \newline
    \textbf{- 3: Good}. The conversation has a challenging/deep/twist topic/discussion. Living the readers to be excited.
    \\
    Coherence & 1–3 & Ensure logical flow between utterances. & Verify that each response directly relates to the previous one. Flag any abrupt topic shifts or irrelevant responses. \newline
    \textbf{- 1: Poor}. The utterances in the dialogue are completely unrelated and nonsensical. \newline
    \textbf{- 2: Fair}. The dialogue is somewhat coherent, but it contains some unrelated utterances. \newline
    \textbf{- 3: Good}. The dialogue is fully coherent, with no hallucinations or inconsistencies in the utterances.
    \\
    Naturalness & 1–3 & Determine how human-like and contextually appropriate the dialogue is. & Look for natural rhythms, idiomatic expressions, and smooth transitions. Penalize mechanical or repetitive phrasing. \newline
    \textbf{- 1: Poor}. The dialogue is completely unnatural and robotic, with awkward phrasing and obvious AI generation. \newline
    \textbf{- 2: Fair}. The dialogue is somewhat natural but still has noticeable AI-like patterns (e.g., repetitive phrasing or awkward transitions). \newline
    \textbf{- 3: Good}. The dialogue feels entirely natural, like a real human conversation, with no clear signs of AI involvement.
    \\
    Cultural Relevance & 0–3 & Assess the accuracy of cultural references. & Identify cultural references and evaluate their correctness within the intended cultural context. \newline
    \textbf{- 0:} The dialogue doesn’t have cultural aspect \newline
    \textbf{- 1: Poor}. The dialogue is entirely irrelevant to the intended culture, containing inaccuracies or stereotypes. \newline
    \textbf{- 2: Fair}. The dialogue has some cultural relevance but includes noticeable inaccuracies or lacks depth. \newline
    \textbf{- 3: Good}. The dialogue is fully culturally correct, accurately reflecting norms, knowledge, and references authentically.
    \\
    Profile Detection & Y/N & Check alignment with provided persona descriptions. & Compare character behavior, tone, and language to the persona descriptions. \textbf{Y}: Traits are clearly represented. \textbf{N}: Traits are absent. \\
    Correctness & Y/N & Verify adherence to the topic constraints. & Confirm that both specified topics from the prompt are clearly addressed. \textbf{Y}: Both topics appear. \textbf{N}: At least one topic is missing. \\
    \bottomrule
  \end{tabular}
  \caption{Dialogue Evaluation Rubrics.}
  \label{tab:dialogue_rubrics}
\end{table*}

\section{Human Annotation Platform}
\label{sec:annotation_platform_appendix}
\begin{figure*}[!htbp]
    \centering
    \small
    \begin{subfigure}[t]{0.48\linewidth}
        \includegraphics[width=\linewidth]{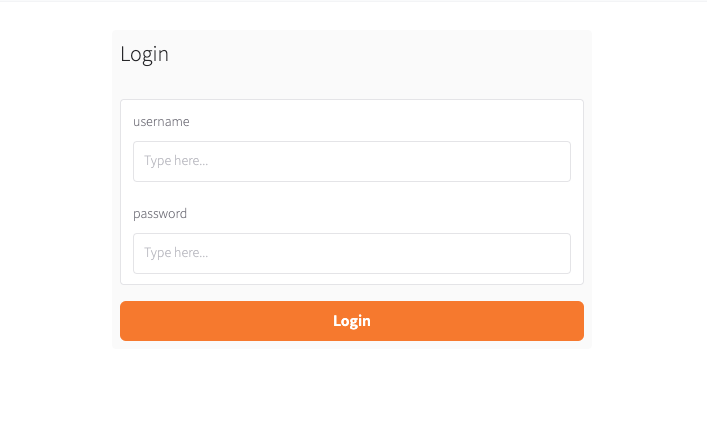}
        \caption{Login Page.}
        \label{fig:annotation-platform-1}
    \end{subfigure}
    \hfill
    \begin{subfigure}[t]{0.48\linewidth}
        \includegraphics[width=\linewidth]{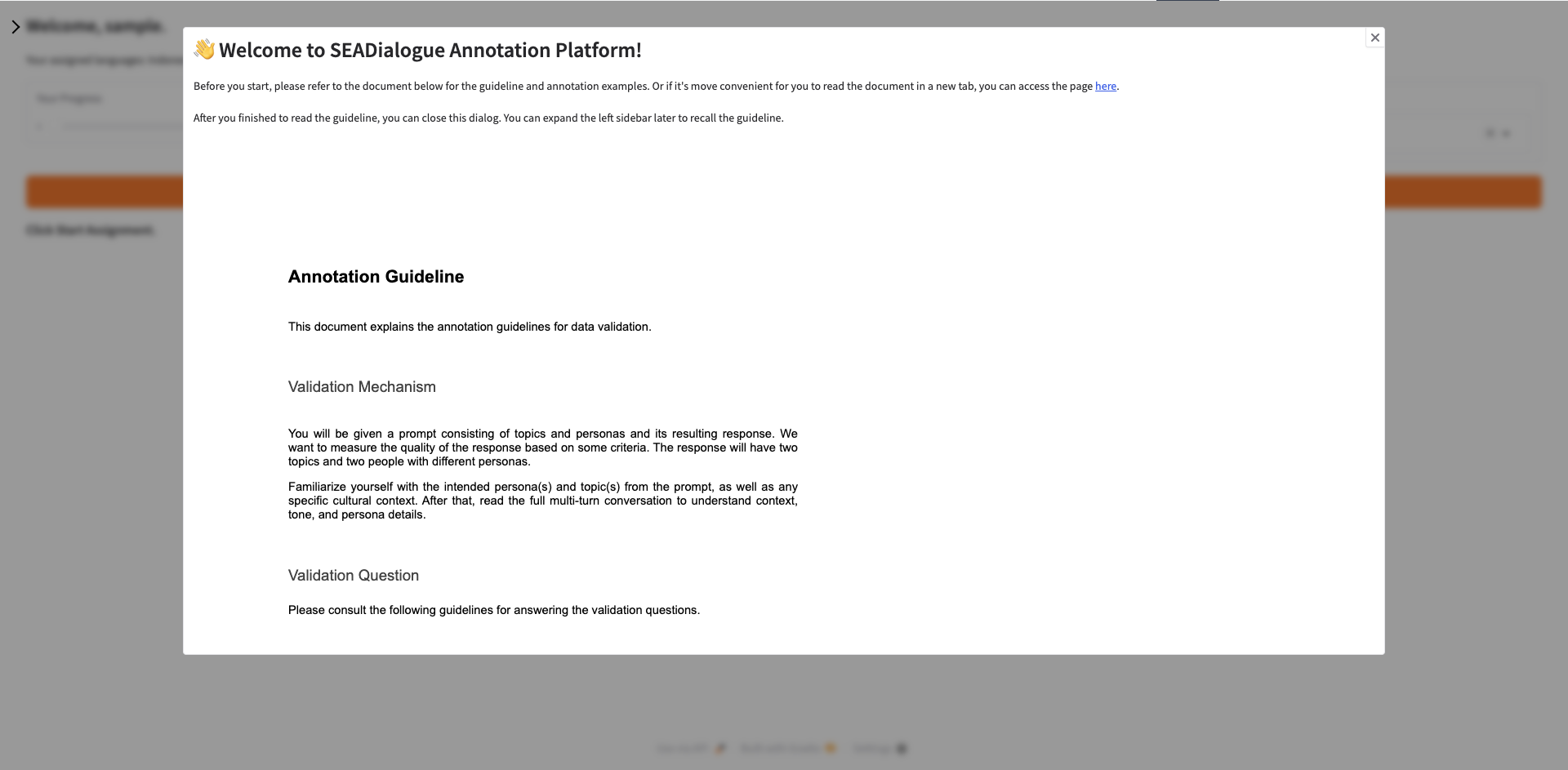}
        \caption{Welcome Page.}
        \label{fig:annotation-platform-2}
    \end{subfigure}

    \vspace{0.5em}

    \begin{subfigure}[t]{0.48\linewidth}
        \includegraphics[width=\linewidth]{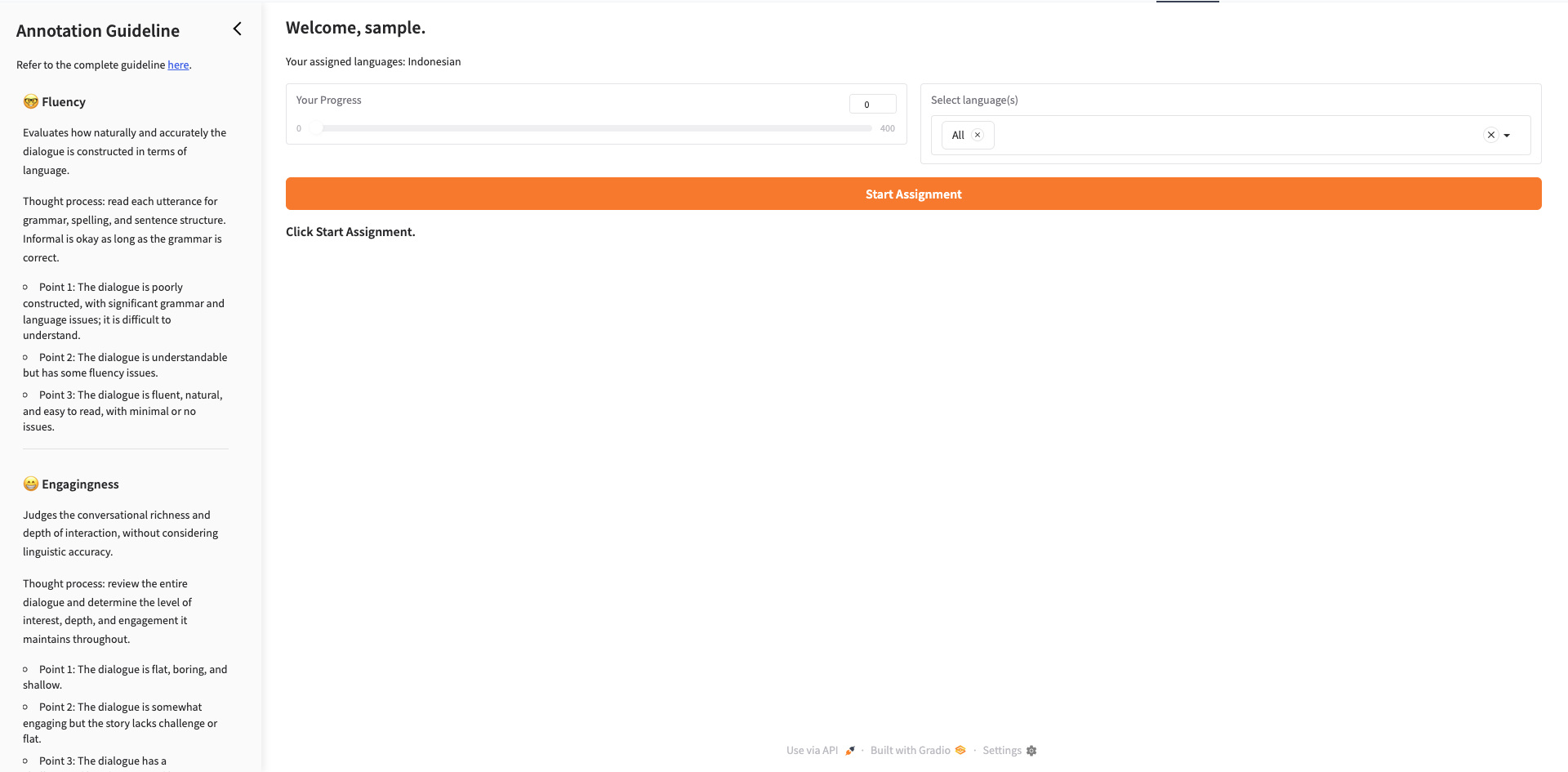}
        \caption{Main Page.}
        \label{fig:annotation-platform-3}
    \end{subfigure}
    \hfill
    \begin{subfigure}[t]{0.48\linewidth}
        \includegraphics[width=\linewidth]{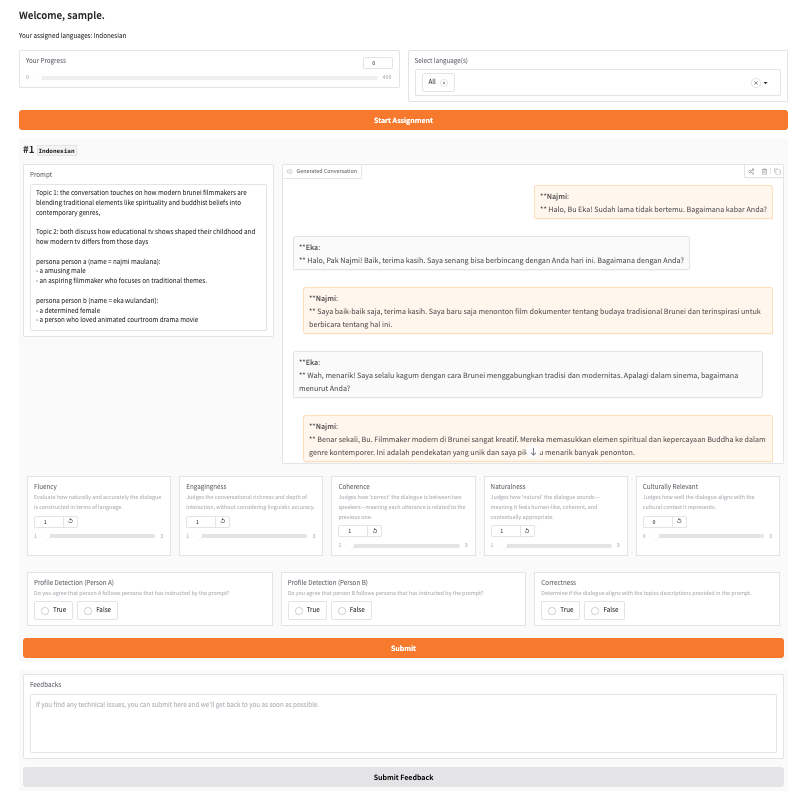}
        \caption{Annotation Task Assignments Page.}
        \label{fig:annotation-platform-4}
    \end{subfigure}

    \caption{\centering Screenshots of the $\datasetname$ Annotation Platform: (a) Login Page, (b) Welcome Page, (c) Main Page, and (d) Annotation Task Assignments Display.}
    \label{fig:annotation-platform}
\end{figure*}

Figures~\ref{fig:annotation-platform-1}--\ref{fig:annotation-platform-4} show the proprietary annotation platform we built to support human annotators in performing their tasks.

\section{Automatic Evaluation}
\label{sec:automatic-evaluation-detail}

\subsection{G-Eval}

We perform G-Eval by modifying the original G-Eval prompt template \cite{liu2023g}. Table \ref{tab:g-eval-prompt-template} shows the one that we use.
\begin{table*}[!htbp]
  \centering
  \begin{tabularx}{\textwidth}{@{}X@{}}
    \toprule
    \textbf{Prompt Template} \\
    \midrule
    You will be given one generated dialogue between two people, each with a distinct persona, discussing two predetermined topics. \\[0.5ex]
    
    Your task is to rate the dialogue on one metric. \\[0.5ex]
    
    Please make sure you read and understand these instructions carefully. Please keep this document open while reviewing, and refer to it as needed. \\[1ex]
    
    \textbf{Evaluation Criteria:} \\
    \{metric's evaluation criteria\} \\[1ex]
    
    \textbf{Evaluation Steps:} \\
    \{metric's evaluation steps\} \\[1ex]
    
    \textbf{Example:} \\[0.5ex]
    
    \textbf{Generated Dialogue:} \\
    \{\{Dialogue\}\} \\[1ex]
    
    \textbf{Evaluation Form (scores ONLY):} \\
    - \{metric's name\}: \\ 
    \bottomrule
  \end{tabularx}
  \caption{Automatic Evaluation G-Eval Prompt Template.}
  \label{tab:g-eval-prompt-template}
\end{table*}
Every metric follows this general template, substituting placeholders with its metric information. The \textit{metric evaluation criteria} correspond to the Evaluation Criteria column in Table \ref{tab:dialogue_rubrics}. For \textit{metric evaluation steps}, each metric has its specific procedures, which are shown in Table \ref{tab:g-eval-steps}.
Additionally, for some metrics, they provide additional information before presenting the dialogue. Language information is added for the \textit{Fluency}, \textit{Naturalness}, and \textit{Culturally Relevance} metrics. Topic descriptions are provided for the \textit{Correctness} metrics. Lastly, Persona descriptions are given for \textit{Profile Detection} metrics. 
\begin{table*}[!ht]
  \centering
  \begin{tabularx}{\textwidth}{@{}X@{}}
    \toprule
    \textbf{Evaluation Steps} \\
    \midrule

    \textit{Fluency:}
    \vspace{-\topsep}
    \begin{enumerate}
      \item Read the entire dialogue thoroughly. \vspace{-\topsep}
      \item Review each utterance for grammatical correctness, spelling, and sentence structure. \vspace{-\topsep}
      \item Informal language is fine, as long as it's grammatically correct. \vspace{-\topsep}
      \item Assign a score for Fluency based on the Evaluation Criteria. \vspace{-\topsep}
    \end{enumerate}

    \vspace{1ex}
    \textit{Engagingness:}
    \vspace{-\topsep}
    \begin{enumerate}
      \item Read the entire dialogue thoroughly. \vspace{-\topsep}
      \item Identify topic depth and richness. \vspace{-\topsep}
      \item Determine the level of interest, depth, and engagement it maintains throughout. \vspace{-\topsep}
      \item Assign a score for Engagingness based on the Evaluation Criteria. \vspace{-\topsep}
    \end{enumerate}

    \vspace{1ex}
    \textit{Coherence:}
    \vspace{-\topsep}
    \begin{enumerate}
      \item Read the entire dialogue thoroughly. \vspace{-\topsep}
      \item Determine whether each utterance logically follows the previous one. \vspace{-\topsep}
      \item Look for any abrupt or irrelevant shifts in the conversation. \vspace{-\topsep}
      \item Assign a score for Coherence based on the Evaluation Criteria. \vspace{-\topsep}
    \end{enumerate}

    \vspace{1ex}
    \textit{Naturalness:}
    \vspace{-\topsep}
    \begin{enumerate}
      \item Read the entire dialogue thoroughly. \vspace{-\topsep}
      \item Assess human-likeness of language. Evaluate if the wording, tone, and sentence structure feel authentic and appropriate for spoken conversation. \vspace{-\topsep}
      \item Identify AI-like artifacts. Look for robotic phrasing, overly formal or generic responses, or repeated sentence patterns. \vspace{-\topsep}
      \item Assign a score for Naturalness based on the Evaluation Criteria. \vspace{-\topsep}
    \end{enumerate}

    \vspace{1ex}
    \textit{Cultural Relevance:}
    \vspace{-\topsep}
    \begin{enumerate}
      \item Read the entire dialogue thoroughly. \vspace{-\topsep}
      \item Determine the presence of any cultural references within the dialogue. \vspace{-\topsep}
      \item Look for cultural references in the dialogue and decide if they are accurate. \vspace{-\topsep}
      \item Assign a score for Culturally Relevance metric based on the Evaluation Criteria. \vspace{-\topsep}
    \end{enumerate}

    \vspace{1ex}
    \textit{Correctness:}
    \vspace{-\topsep}
    \begin{enumerate}
      \item Identify the two predetermined topics. \vspace{-\topsep}
      \item Read the entire dialogue carefully. \vspace{-\topsep}
      \item Go through each utterance in the dialogue and see if it fits these topics. \vspace{-\topsep}
      \item Assign a score for Correctness based on the Evaluation Criteria. \vspace{-\topsep}
    \end{enumerate}

    \textit{Profile Detection:}
    \vspace{-\topsep}
    \begin{enumerate}
      \item Read the Persona descriptions for Person A and Person B carefully. \vspace{-\topsep}
      \item Read the entire dialogue attentively. \vspace{-\topsep}
      \item Identify whether the dialogue includes explicit or implicit elements that showcase each speaker's persona. \vspace{-\topsep}
      \item Assign score for Profile Detection based on the Evaluation Criteria for each person. \vspace{-\topsep}
    \end{enumerate}
    \\
    \bottomrule
  \end{tabularx}
  \caption{Evaluation steps for each metric using G-EVAL.}
  \label{tab:g-eval-steps}
\end{table*}

For M-Prometheus, we run the model using a prompt similar to the one presented in the paper \cite{pombal2025mprometheussuiteopenmultilingual}. Table \ref{tab:m-prometeus-prompt-template} shows the prompt template.
\begin{table*}[!ht]
  \centering
  \begin{tabularx}{\textwidth}{@{}X@{}}
    \toprule
    \textbf{Prompt Template} \\
    \midrule
    \#\#\#Task Description: An instruction (might include an Input inside it), a response to evaluate, and a score rubric representing a evaluation criteria are given. \\
     1. Write a detailed feedback that assess the quality of the response strictly based on the given score rubric, not evaluating in general. \\
     2. After writing a feedback, write a score that is an integer between \{min\_score\} and \{max\_score\}. You should refer to the score rubric. \\
     3. The output format should look as follows: "Feedback: (write a feedback for criteria) [RESULT] (an integer number between \{min\_score\} and \{max\_score\})" \\
     4. Please do not generate any other opening, closing, and explanations. \\
    
     \#\#\#The instruction to evaluate:
        \{instruction\}
     The original prompt for the dialogue was: \{dialogue\_prompt\} \\
    
     \#\#\#Response to evaluate:
        \{generated\_dialogue\} \\
    
     \#\#\#Score Rubrics:
        \{score\_rubrics\} \\
    
     \#\#\#Feedback: \\
    \bottomrule
  \end{tabularx}
  \caption{Automatic Evaluation M-Prometheus Prompt Template.}
  \label{tab:m-prometeus-prompt-template}
\end{table*}
The \textit{instruction} placeholder is replaced by the first sentence from the Evaluation Criteria column in Table \ref{tab:dialogue_rubrics}, while \textit{score\_rubrics} is substituted with the description of each score from the same column.

For the R3 model, we employ the evaluation using the model \cite{anugraha2025r3robustrubricagnosticreward} following the pointwise evaluation prompt template from the paper. Table \ref{tab:R3-prompt-template} displays the template of the prompt.
\begin{table*}[!ht]
  \centering
  \begin{tabularx}{\textwidth}{@{}X@{}}
    \toprule
    \textbf{Prompt Template} \\
    \midrule
    Evaluate the response based on the given task, input, response, and evaluation rubric. Provide a fair and detailed assessment following the rubric. \\

     \#\#\# TASK \\
        \{instruction\} \\
    
     \#\#\# INPUT \\
        \{dialogue\_prompt\} \\
    
     \#\#\# RESPONSE \\
        \{generated\_dialogue\} \\
    
     \#\#\# EVALUATION RUBRIC \\
        \{score\_rubrics\} \\
    
     \#\#\# OUTPUT FORMAT \\
     Return a JSON response in the following format: \\
    \{\{ \\
     "explanation": "Explanation of why the response received a particular score",
     "score": "Score assigned to the response based on the rubric between \{score\_range\}" \\
    \}\} \\
    
     \#\#\# EVALUATION \\
    \bottomrule
  \end{tabularx}
  \caption{Automatic Evaluation R3 Prompt Template.}
  \label{tab:R3-prompt-template}
\end{table*}
We replace these placeholders in the same way as described previously for the M-Prometheus prompt.

\section{Automatic Evaluation Results}
This section presents the full results of the automatic evaluation using M-Prometheus \cite{pombal2025mprometheussuiteopenmultilingual} and the R3 \cite{anugraha2025r3robustrubricagnosticreward} reward model on our dataset. Table~\ref{tab:automatic-metrics-m-prometheus}  displays the M-Prometheus score results for each metric, while Table \ref{tab:automatic-metrics-r3} shows the R3 score results for every metric.

\begin{table*}[!ht]
    \centering
    \resizebox{0.8\textwidth}{!}{
    \begin{tabular}{l|c c c c }
    \toprule
    \textbf{Model} & \texttt{ind} & \texttt{jav} & \texttt{min} & \texttt{tha} \\
    \midrule
    \textbf{Coherence} & & & &  \\
    \quad Aya-8B-Expanse & $3.00 \pm 0.00$ & $3.00 \pm 0.00$ & $3.00 \pm 0.00$ & $2.80 \pm 0.53$ \\
    \quad Gemini 1.5 Flash & $3.00 \pm 0.00$ & $3.00 \pm 0.00$ & $3.00 \pm 0.00$ & $3.00 \pm 0.00$ \\
    \quad GPT-4o mini & $3.00 \pm 0.00$ & $3.00 \pm 0.00$ & $3.00 \pm 0.00$ & $3.00 \pm 0.00$ \\
    \quad Llama-3.1-Instruct & $3.00 \pm 0.00$ & $2.42 \pm 0.82$ & $2.79 \pm 0.57$  & $2.92 \pm 0.31$  \\
    \midrule
    \textbf{Culturally Relevance}  & & & & \\
    \quad Aya-8B-Expanse & $3.00 \pm 0.00$ & $3.00 \pm 0.00$ & $3.00 \pm 0.00$  & $2.54 \pm 0.88$  \\
    \quad Gemini 1.5 Flash & $3.00 \pm 0.00$ & $3.00 \pm 0.00$ & $3.00 \pm 0.00$  & $3.00 \pm 0.00$  \\
    \quad GPT-4o mini & $3.00 \pm 0.00$ & $3.00 \pm 0.00$ & $2.99 \pm 0.10$  & $3.00 \pm 0.00$  \\
    \quad Llama-3.1-Instruct & $2.96 \pm 0.20$ & $2.01 \pm 1.08$ & $2.41 \pm 0.83$  & $2.66 \pm 0.52$  \\
    \midrule
    \textbf{Engagingness}  & & & & \\
    \quad Aya-8B-Expanse & $3.00 \pm 0.00$ & $3.00 \pm 0.00$ & $3.00 \pm 0.00$  & $2.46 \pm 0.63$  \\
    \quad Gemini 1.5 Flash & $3.00 \pm 0.00$ & $2.99 \pm 0.10$ & $2.98 \pm 0.14$  & $2.96 \pm 0.20$  \\
    \quad GPT-4o mini & $3.00 \pm 0.00$ & $3.00 \pm 0.00$ & $3.00 \pm 0.00$  & $2.97 \pm 0.17$  \\
    \quad Llama-3.1-Instruct & $2.68 \pm 0.49$ & $1.68 \pm 0.65$ & $2.02 \pm 0.72$  & $2.07 \pm 0.46$  \\
    \midrule
    \textbf{Fluency}  & & & & \\
    \quad Aya-8B-Expanse & $3.00 \pm 0.00$ & $3.00 \pm 0.00$ & $3.00 \pm 0.00$  & $2.55 \pm 0.81$  \\
    \quad Gemini 1.5 Flash & $3.00 \pm 0.00$ & $3.00 \pm 0.00$ & $3.00 \pm 0.00$  & $3.00 \pm 0.00$  \\
    \quad GPT-4o mini & $3.00 \pm 0.00$ & $3.00 \pm 0.00$ & $3.00 \pm 0.00$  & $3.00 \pm 0.00$  \\
    \quad Llama-3.1-Instruct & $3.00 \pm 0.00$ & $2.17 \pm 0.87$ & $2.53 \pm 0.70$  & $2.69 \pm 0.61$  \\
    \midrule
    \textbf{Naturalness}  & & & & \\
    \quad Aya-8B-Expanse & $3.00 \pm 0.00$ & $2.99 \pm 0.10$ & $2.99 \pm 0.10$  & $2.45 \pm 0.78$  \\
    \quad Gemini 1.5 Flash & $2.99 \pm 0.10$ & $3.00 \pm 0.00$ & $2.95 \pm 0.26$  & $2.99 \pm 0.10$  \\
    \quad GPT-4o mini & $2.96 \pm 0.28$ & $2.99 \pm 0.10$ & $3.00 \pm 0.00$  & $3.00 \pm 0.00$  \\
    \quad Llama-3.1-Instruct & $2.94 \pm 0.24$ & $1.95 \pm 0.67$ & $2.24 \pm 0.67$  & $2.34 \pm 0.65$  \\
    \midrule
    \textbf{Correctness}  & & & & \\
    \quad Aya-8B-Expanse & $1.00 \pm 0.00$ & $1.00 \pm 0.00$ & $1.00 \pm 0.00$   & $0.88 \pm 0.33$   \\
    \quad Gemini 1.5 Flash & $1.00 \pm 0.00$ & $1.00 \pm 0.00$ & $1.00 \pm 0.00$   & $0.99 \pm 0.10$   \\
    \quad GPT-4o mini & $1.00 \pm 0.00$ & $1.00 \pm 0.00$ & $1.00 \pm 0.00$   & $1.00 \pm 0.00$   \\
    \quad Llama-3.1-Instruct & $0.98 \pm 0.14$ & $0.77 \pm 0.42$ & $0.79 \pm 0.41$   & $0.88 \pm 0.33$   \\
    \midrule
    \textbf{Profile Detection}  & & & & \\
    \quad Aya-8B-Expanse & $0.98 \pm 0.14$ & $0.99 \pm 0.07$ & $0.99 \pm 0.07$   & $0.79 \pm 0.41$   \\
    \quad Gemini 1.5 Flash & $0.99 \pm 0.10$ & $1.00 \pm 0.00$ & $1.00 \pm 0.00$   & $0.98 \pm 0.12$   \\
    \quad GPT-4o mini & $1.00 \pm 0.00$ & $1.00 \pm 0.00$ & $0.99 \pm 0.07$   & $0.98 \pm 0.14$   \\
    \quad Llama-3.1-Instruct & $0.98 \pm 0.14$ & $0.83 \pm 0.38$ & $0.95 \pm 0.22$   & $0.92 \pm 0.27$   \\
    \bottomrule
    \end{tabular}
    }
    \caption{M-Prometheus results on model's conversational generation and instruction following capabilities.}
    \label{tab:automatic-metrics-m-prometheus}
\end{table*}


\begin{table*}[!ht]
    \centering
    \resizebox{0.8\textwidth}{!}{
    \begin{tabular}{l|c c c c}
    \toprule
    \textbf{Model} & \texttt{ind} & \texttt{jav} & \texttt{min} & \texttt{tha} \\
    \midrule
    \textbf{Coherence} & & & & \\
    \quad Aya-8B-Expanse & $2.94 \pm 0.24$ & $2.89 \pm 0.31$ & $2.94 \pm 0.24$ &  $1.85 \pm 0.36$  \\
    \quad Gemini 1.5 Flash & $3.00 \pm 0.00$ & $2.97 \pm 0.17$ & $2.98 \pm 0.14$  & $2.96 \pm 0.20$  \\
    \quad GPT-4o mini & $2.99 \pm 0.10$ & $2.99 \pm 0.10$ & $2.98 \pm 0.14$  & $2.96 \pm 0.20$  \\
    \quad Llama-3.1-Instruct & $2.83 \pm 0.38$ & $2.05 \pm 0.54$ & $2.53 \pm 0.58$  & $2.68 \pm 0.47$  \\
    \midrule
    \textbf{Culturally Relevance} & & & & \\
    \quad Aya-8B-Expanse & $2.67 \pm 0.49$ & $2.74 \pm 0.46$ & $2.65 \pm 0.59$  & $1.71 \pm 0.50$  \\
    \quad Gemini 1.5 Flash & $2.83 \pm 0.38$ & $2.96 \pm 0.20$ & $2.92 \pm 0.27$  & $2.79 \pm 0.41$  \\
    \quad GPT-4o mini & $2.71 \pm 0.46$ & $2.93 \pm 0.26$ & $2.69 \pm 0.54$  & $2.73 \pm 0.47$  \\
    \quad Llama-3.1-Instruct & $2.51 \pm 0.52$ & $1.94 \pm 0.55$ & $2.26 \pm 0.61$  & $2.22 \pm 0.48$  \\
    \midrule
    \textbf{Engagingness}   & & & & \\
    \quad Aya-8B-Expanse & $2.77 \pm 0.42$ & $2.71 \pm 0.46$ & $2.76 \pm 0.43$  & $1.48 \pm 0.50$  \\
    \quad Gemini 1.5 Flash & $2.76 \pm 0.43$ & $2.61 \pm 0.49$ & $2.64 \pm 0.48$  & $2.69 \pm 0.46$  \\
    \quad GPT-4o mini & $2.75 \pm 0.44$ & $2.67 \pm 0.47$ & $2.54 \pm 0.50$  & $2.60 \pm 0.49$  \\
    \quad Llama-3.1-Instruct & $2.17 \pm 0.38$ & $1.60 \pm 0.53$ & $1.93 \pm 0.43$  & $2.01 \pm 0.27$  \\
    \midrule
    \textbf{Fluency}   & & & & \\
    \quad Aya-8B-Expanse & $2.92 \pm 0.27$ & $2.85 \pm 0.36$ & $2.80 \pm 0.43$  & $1.40 \pm 0.49$  \\
    \quad Gemini 1.5 Flash & $2.98 \pm 0.14$ & $2.96 \pm 0.20$ & $2.93 \pm 0.26$  & $2.92 \pm 0.27$  \\
    \quad GPT-4o mini & $2.92 \pm 0.27$ & $2.95 \pm 0.22$ & $2.83 \pm 0.43$  & $2.97 \pm 0.17$  \\
    \quad Llama-3.1-Instruct & $2.76 \pm 0.43$ & $1.86 \pm 0.53$ & $2.18 \pm 0.54$  & $2.30 \pm 0.46$  \\
    \midrule
    \textbf{Naturalness}   & & & & \\
    \quad Aya-8B-Expanse & $2.81 \pm 0.39$ & $2.85 \pm 0.36$ & $2.72 \pm 0.45$  & $1.58 \pm 0.50$  \\
    \quad Gemini 1.5 Flash & $2.97 \pm 0.17$ & $2.83 \pm 0.38$ & $2.87 \pm 0.34$  & $2.73 \pm 0.45$  \\
    \quad GPT-4o mini & $2.89 \pm 0.31$ & $2.81 \pm 0.39$ & $2.80 \pm 0.40$  & $2.78 \pm 0.42$  \\
    \quad Llama-3.1-Instruct & $2.29 \pm 0.46$ & $1.85 \pm 0.44$ & $2.04 \pm 0.37$  & $2.08 \pm 0.27$  \\
    \textbf{Correctness}   & & & & \\
    \quad Aya-8B-Expanse & $0.90 \pm 0.30$ & $0.92 \pm 0.27$ & $0.95 \pm 0.22$  & $0.13 \pm 0.34$  \\
    \quad Gemini 1.5 Flash & $1.00 \pm 0.00$ & $1.00 \pm 0.00$ & $1.00 \pm 0.00$  & $1.00 \pm 0.00$  \\
    \quad GPT-4o mini & $1.00 \pm 0.00$ & $1.00 \pm 0.00$ & $0.99 \pm 0.10$  & $1.00 \pm 0.00$  \\
    \quad Llama-3.1-Instruct & $0.89 \pm 0.31$ & $0.55 \pm 0.50$ & $0.75 \pm 0.44$  & $0.73 \pm 0.45$  \\
    \midrule
    \textbf{Profile Detection}   & & & & \\
    \quad Aya-8B-Expanse & $0.89 \pm 0.32$ & $0.88 \pm 0.33$ & $0.92 \pm 0.27$  & $0.31 \pm 0.46$  \\
    \quad Gemini 1.5 Flash & $0.85 \pm 0.36$ & $0.88 \pm 0.33$ & $0.86 \pm 0.35$  & $0.80 \pm 0.40$  \\
    \quad GPT-4o mini & $0.84 \pm 0.37$ & $0.88 \pm 0.33$ & $0.87 \pm 0.34$  & $0.79 \pm 0.41$  \\
    \quad Llama-3.1-Instruct & $0.76 \pm 0.43$ & $0.41 \pm 0.49$ & $0.66 \pm 0.47$  & $0.66 \pm 0.47$  \\
    \bottomrule
    \end{tabular}
    }
    \caption{R3 results on model's conversational generation and instruction following capabilities.}
    \label{tab:automatic-metrics-r3}
\end{table*}


\section{Human-Automatic Evaluation Correlation}

\begin{figure}[p]
    \centering
    \includegraphics[width=\linewidth]{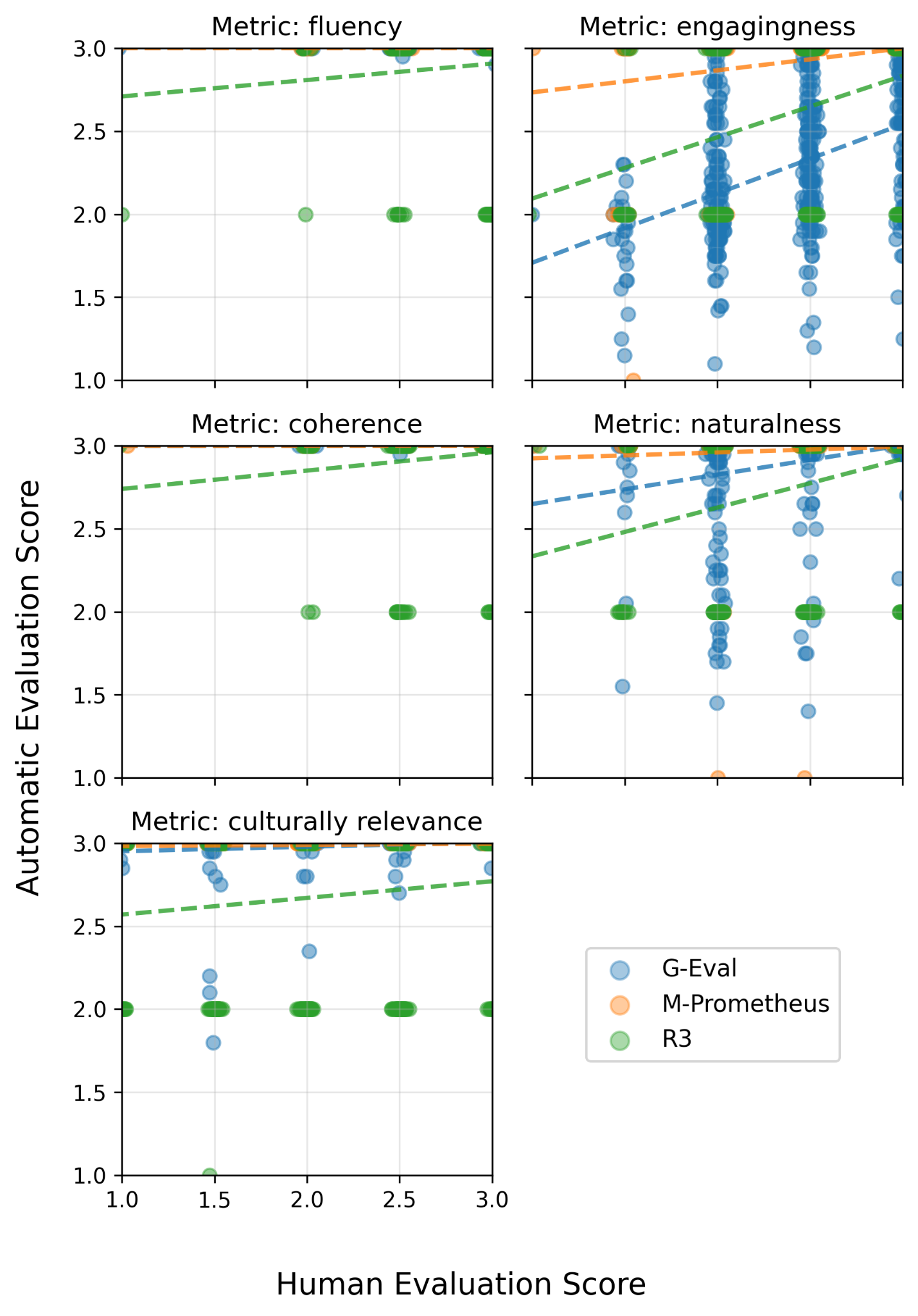}
    \caption{Visualization of the Correlation Between Automatic Evaluation and Human Annotations for All Generation Capability Metrics on Indonesia.}
    \label{fig:corr-id-all}
\end{figure}

\begin{figure}[p]
    \centering
    \includegraphics[width=\linewidth]{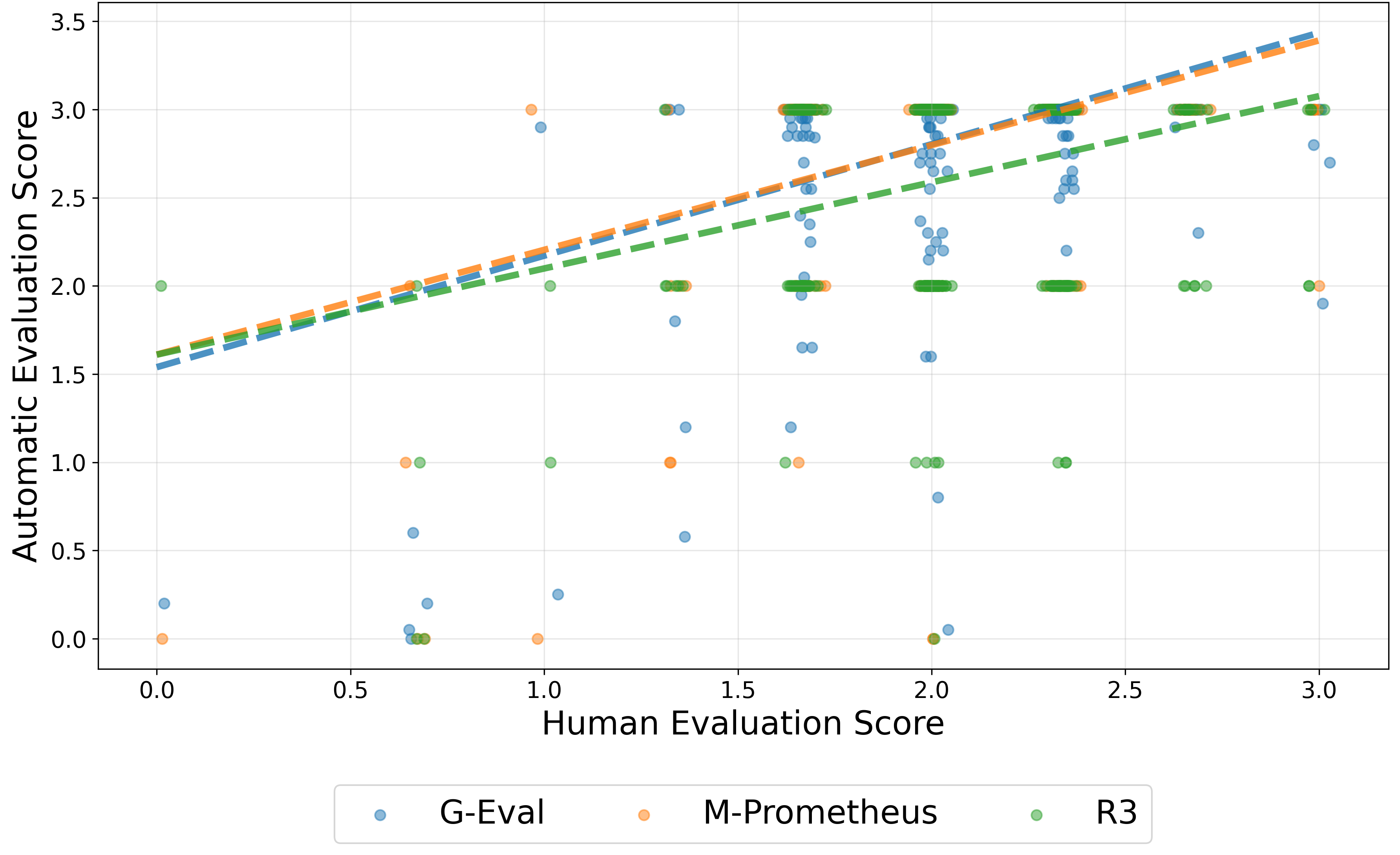}
    \caption{Visualization of the Correlation Between Automatic Evaluation and Human Annotations for Culturally Relevance Metrics on Minangkabau.}
    \label{fig:corr-min-culture}
\end{figure}

\begin{figure}[p]
    \centering
    \includegraphics[width=\linewidth]{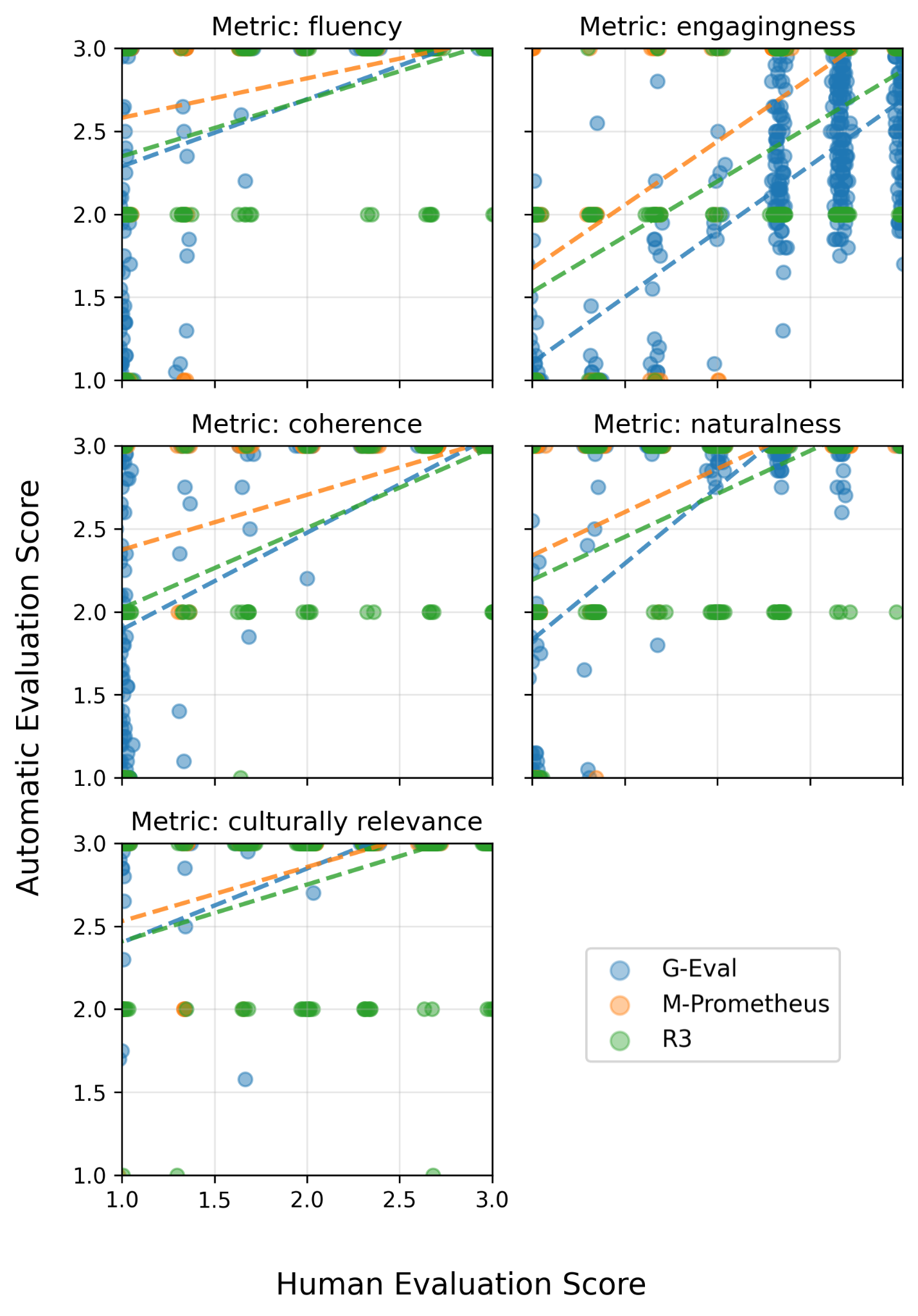}
    \caption{Visualization of the Correlation Between Automatic Evaluation and Human Annotations for All Generation Capability Metrics on Javanese.}
    \label{fig:corr-jav-all}
\end{figure}

\begin{figure}[p]
    \centering
    \includegraphics[width=\linewidth]{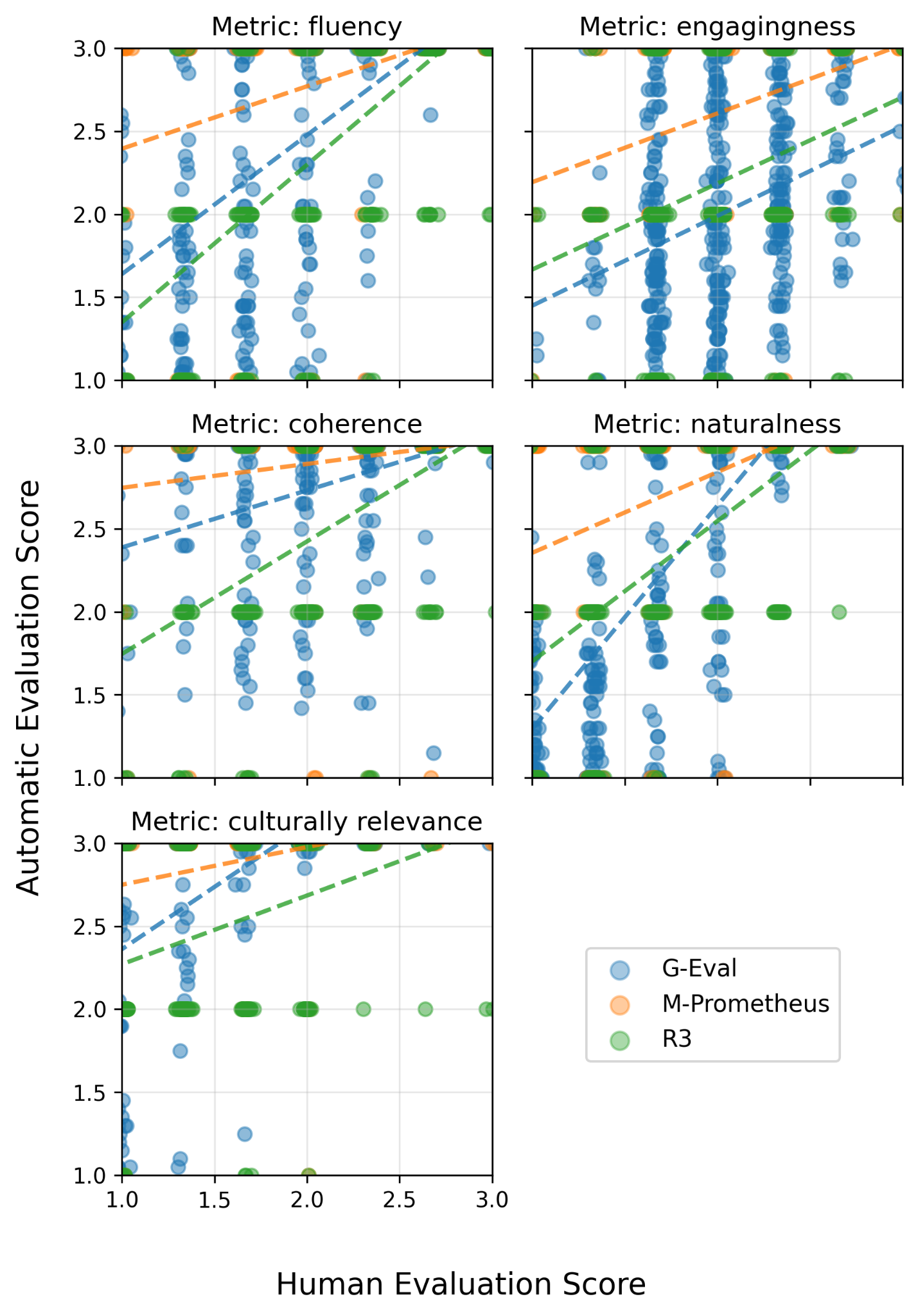}
    \caption{Visualization of the Correlation Between Automatic Evaluation and Human Annotations for All Generation Capability Metrics on Thai.}
    \label{fig:corr-thai-all}
\end{figure}

This section presents the supplementary figures and detailed correlation data derived from the human evaluation alongside LLM-as-judge assessments. We extend our alignment analysis to instruction-following capabilities by binarizing human scores for the Correctness and Profile Detection metrics, which are then used as labels. The corresponding scores from the automatic evaluation serve as predictions. Based on these predictions, we compute precision, recall, F1 score, and accuracy to evaluate the reliability of the automatic evaluation. These results are reported in Tables \ref{tab:auto_human_instruc_id}, \ref{tab:auto_human_instruc}, \ref{tab:auto_human_instruc_jav}, and \ref{tab:auto_human_instruc_thai}. Figures \ref{fig:corr-id-all}–\ref{fig:corr-thai-all} further illustrate the correlation between automatic and human evaluations across the Indonesian, Minangkabau, Javanese, and Thai subsets. Tables \ref{tab:auto_human_correlations_id}, \ref{tab:auto_human_correlations_jav}, and \ref{tab:auto_human_correlations_thai} report the corresponding quantitative alignment results.

\begin{table*}[!ht]
    \centering
    \resizebox{0.6\textwidth}{!}{
    \begin{tabular}{l|c c c}
    \toprule
    \textbf{} & \textbf{G-Eval} & \textbf{M-Prometheus} & \textbf{R3} \\

    \midrule
    \textit{Pearson} \\
    \quad Coherence & 0.0555 & NaN & \textbf{0.1336} \\
    \quad Culturally Relevance & \textbf{0.1704} & 0.0489 & 0.1508 \\
    \quad Engagingness & \textbf{0.4282} & 0.2093 & 0.3353 \\
    \quad Fluency & 0.0123 & NaN & \textbf{0.0795} \\
    \quad Naturalness & 0.2429 & 0.0733 & \textbf{0.2658} \\

    \midrule
    \textit{Spearman} \\
    \quad Coherence & 0.0684 & NaN & \textbf{0.1440}\\
    \quad Culturally Relevance & \textbf{0.1846} & 0.0587 & 0.1587\\
    \quad Engagingness & \textbf{0.4333} & 0.1841 & 0.3249 \\
    \quad Fluency & \textbf{0.0460} & NaN & 0.0397 \\
    \quad Naturalness & \textbf{0.3188} & 0.0835 & 0.2690 \\

    \midrule
    \textit{KendallTau} \\
    \quad Coherence & 0.0673 & NaN & \textbf{0.1417} \\
    \quad Culturally Relevance & \textbf{0.1633} & 0.0527 & 0.1424 \\
    \quad Engagingness & \textbf{0.3440} & 0.1711 & 0.3020 \\
    \quad Fluency & \textbf{0.0456} & NaN & 0.0394 \\
    \quad Naturalness & \textbf{0.2835} & 0.0783 & 0.2529 \\

    \bottomrule
    \end{tabular}}
    \caption{Automatic Evaluations and Human Annotations Correlations on Indonesian. M-Prometheus assigns the same values to both Coherence and Fluency, making correlation calculation not possible (resulting in NaN).}
    \label{tab:auto_human_correlations_id}
\end{table*}
\begin{table*}[!ht]
    \centering
    \resizebox{0.55\textwidth}{!}{
    \begin{tabular}{l|c c c c}
    \toprule
    \textbf{LLM Judge} & \textbf{P} & \textbf{R} & \textbf{F1} & \textbf{Accuracy}\\

    \midrule
    \textit{G-Eval} \\
    \quad Correctness & 0.91 & \textbf{0.99} & 0.94 & \textbf{0.91} \\
    \quad Profile Detection & 0.41 & \textbf{0.99} & 0.58 & 0.42 \\

    \midrule
    \textit{M-Prometheus} \\
    \quad Correctness & 0.91 & 0.99 & \textbf{0.95} & 0.91 \\
    \quad Profile Detection & 0.41 & 0.98 & 0.57 & 0.41\\

    \midrule
    \textit{R3} \\
    \quad Correctness & \textbf{0.92} & 0.96 & 0.94 & 0.90 \\
    \quad Profile Detection & \textbf{0.46} & 0.94 & \textbf{0.61} & \textbf{0.52}\\

    \bottomrule
    \end{tabular}}
    \caption{\centering Comparison of Automatic Evaluation Predictions to Human Annotations for Instruction Following Metrics on Indonesian.}
    \label{tab:auto_human_instruc_id}
\end{table*}
\begin{table*}[!ht]
    \centering
    \resizebox{0.49\textwidth}{!}{
    \begin{tabular}{l|c c c c}
    \toprule
    \textbf{LLM Judge} & \textbf{P} & \textbf{R} & \textbf{F1} & \textbf{Accuracy}\\
    \midrule
    \textbf{G-Eval} \\
    \quad Correctness & 0.985 & \textbf{0.987} & \textbf{0.986} & \textbf{0.972}  \\
    \quad Profile Detection & 0.977 & 0.966 & 0.971 & 0.945 \\
    \midrule
    \textbf{M-Prometheus} \\
    \quad Correctness & \textbf{0.986} & 0.956 & 0.971 & 0.945  \\
    \quad Profile Detection & 0.973 & \textbf{0.990} & \textbf{0.982} & \textbf{0.965} \\
    \midrule
    \textbf{R3} \\
    \quad Correctness & 0.983 & 0.928 & 0.955 & 0.915  \\
    \quad Profile Detection & \textbf{0.981} & 0.837 & 0.903 & 0.828 \\
    
    \bottomrule
    \end{tabular}
    }
    \caption{\centering Comparison of Automatic Evaluation Predictions to Human Annotations for Instruction Following Metrics on Minangkabau.}
    \label{tab:auto_human_instruc}
    \vspace{-3mm}
\end{table*}

\begin{table*}[!ht]
    \centering
    \resizebox{0.6\textwidth}{!}{
    \begin{tabular}{l|c c c}
    \toprule
    \textbf{} & \textbf{G-Eval} & \textbf{M-Prometheus} & \textbf{R3} \\
    
    \midrule
    \textit{Pearson} \\
    \quad Coherence & \textbf{0.7899} & 0.5496 & 0.7794  \\
    \quad Culturally Relevance & \textbf{0.6228} & 0.5557 & 0.5956 \\
    \quad Engagingness & \textbf{0.8338} & 0.8377 & 0.6935 \\
    \quad Fluency & 0.5007 & 0.3743 & \textbf{0.5105} \\
    \quad Naturalness & \textbf{0.6811} & 0.5668 & 0.5424 \\

    \midrule
    \textit{Spearman} \\
    \quad Coherence & \textbf{0.7846} & 0.4864 & 0.7040  \\
    \quad Culturally Relevance & \textbf{0.6893} & 0.5551 & 0.5956 \\
    \quad Engagingness & 0.6684 & \textbf{0.7180} & 0.5878 \\
    \quad Fluency & 0.5022 & 0.3794 & \textbf{0.5036} \\
    \quad Naturalness & 0.5979 & \textbf{0.6319} & 0.5731 \\

    \midrule
    \textit{KendallTau} \\
    \quad Coherence & \textbf{0.7022} & 0.4505 & 0.6518  \\
    \quad Culturally Relevance & \textbf{0.5775} & 0.4789 & 0.4868 \\
    \quad Engagingness & 0.5372 & \textbf{0.6364} & 0.5235 \\
    \quad Fluency & 0.4426 & 0.3415 & \textbf{0.4535} \\
    \quad Naturalness & 0.4867 & \textbf{0.5483} & 0.4983 \\

    \bottomrule
    \end{tabular}}
    \caption{Automatic Evaluations and Human Annotations Correlations on Javanese.}
    \label{tab:auto_human_correlations_jav}
\end{table*}
\begin{table*}[!ht]
    \centering
    \resizebox{0.55\textwidth}{!}{
    \begin{tabular}{l|c c c c}
    \toprule
    \textbf{LLM Judge} & \textbf{P} & \textbf{R} & \textbf{F1} & \textbf{Accuracy}\\
    
    \midrule
    \textit{G-Eval} \\
    \quad Correctness & 0.846 & \textbf{0.996} & 0.916 & 0.853  \\
    \quad Profile Detection & 0.772 & \textbf{1.000} & \textbf{0.871} & 0.795 \\

    \midrule
    \textit{M-Prometheus} \\
    \quad Correctness & 0.846 & 0.991 & 0.912 & 0.847  \\
    \quad Profile Detection & 0.724 & 0.998 & 0.839 & 0.734 \\

    \midrule
    \textit{R3} \\
    \quad Correctness & \textbf{0.893} & 0.966 & \textbf{0.928} & \textbf{0.880}  \\
    \quad Profile Detection & \textbf{0.822} & 0.903 & 0.861 & \textbf{0.797} \\
    
    \bottomrule
    \end{tabular}}
    \caption{\centering Comparison of Automatic Evaluation Predictions to Human Annotations for Instruction Following Metrics on Javanese.}
    \label{tab:auto_human_instruc_jav}
\end{table*}
\begin{table*}[!ht]
    \centering
    \resizebox{0.6\textwidth}{!}{
    \begin{tabular}{l|c c c}
    \toprule
    \textbf{} & \textbf{G-Eval} & \textbf{M-Prometheus} & \textbf{R3} \\

    \midrule
    \textit{Pearson} \\
    \quad Coherence & 0.4191 & 0.2116 & \textbf{0.5655} \\
    \quad Culturally Relevance & \textbf{0.4994} & 0.2293 & 0.3596 \\
    \quad Engagingness & \textbf{0.3353} & 0.2694 & 0.2842 \\
    \quad Fluency & 0.6591 & 0.3751 & \textbf{0.6986} \\
    \quad Naturalness & \textbf{0.7408} & 0.3751 & 0.5987 \\

    \midrule
    \textit{Spearman} \\
    \quad Coherence & 0.5079 & 0.2092 & \textbf{0.5482} \\
    \quad Culturally Relevance & \textbf{0.5146} & 0.2168 & 0.3798 \\
    \quad Engagingness & \textbf{0.3323} & 0.2657 & 0.2805 \\
    \quad Fluency & 0.6995 & 0.3598 & \textbf{0.7001} \\
    \quad Naturalness & \textbf{0.7251} & 0.3851 & 0.5952 \\

    \midrule
    \textit{KendallTau} \\
    \quad Coherence & 0.4279 & 0.1859 & \textbf{0.4877} \\
    \quad Culturally Relevance & \textbf{0.4214} & 0.1889 & 0.3267 \\
    \quad Engagingness & \textbf{0.2519} & 0.2395 & 0.2485 \\
    \quad Fluency & 0.5778 & 0.3154 & \textbf{0.6103} \\
    \quad Naturalness & \textbf{0.5941} & 0.3388 & 0.5221 \\

    \bottomrule
    \end{tabular}}
    \caption{Automatic Evaluations and Human Annotations Correlations on Thai.}
    \label{tab:auto_human_correlations_thai}
\end{table*}
\begin{table*}[!ht]
    \centering
    \resizebox{0.55\textwidth}{!}{
    \begin{tabular}{l|c c c c}
    \toprule
    \textbf{LLM Judge} & \textbf{P} & \textbf{R} & \textbf{F1} & \textbf{Accuracy}\\

    \midrule
    \textit{G-Eval} \\
    \quad Correctness & 0.8630 & 0.9642 & \textbf{0.9108} & 0.8550 \\
    \quad Profile Detection & 0.5417 & 0.9487 & 0.6897 & 0.5838 \\

    \midrule
    \textit{M-Prometheus} \\
    \quad Correctness & 0.7973 & \textbf{0.9739} & 0.8768 & 0.7900 \\
    \quad Profile Detection & 0.5109 & \textbf{0.9615} & 0.6673 & 0.5325 \\

    \midrule
    \textit{R3} \\
    \quad Correctness & \textbf{0.9406} & 0.8762 & 0.9073 & \textbf{0.8625} \\
    \quad Profile Detection & \textbf{0.6169} & 0.8051 & \textbf{0.6986} & \textbf{0.6608} \\

    \bottomrule
    \end{tabular}}
    \caption{\centering Comparison of Automatic Evaluation Predictions to Human Annotations for Instruction Following Metrics on Thai.}
    \label{tab:auto_human_instruc_thai}
\end{table*}

\end{document}